\documentclass[10pt,twocolumn,letterpaper]{article}

\newif\ifarxiv
\arxivtrue

\ifarxiv    \usepackage[pagenumbers]{cvpr} 
\else       \usepackage{cvpr}              
\fi

\usepackage[accsupp]{axessibility}  
\usepackage{graphicx}
\usepackage{amsmath}
\usepackage{amssymb}
\usepackage{booktabs}

\usepackage{caption}
\usepackage{subcaption}
\usepackage[dvipsnames]{xcolor}

\usepackage{tabu}
\usepackage{tabularx}

\usepackage{multirow}

\usepackage{pifont}
\newcommand{\xmark}{\ding{55}}%

\usepackage[pagebackref,breaklinks,colorlinks]{hyperref}

\usepackage[capitalize]{cleveref}
\crefname{section}{Sec.}{Secs.}
\Crefname{section}{Section}{Sections}
\Crefname{table}{Table}{Tables}
\crefname{table}{Tab.}{Tabs.}

\def\cvprPaperID{7211} 
\def\confName{CVPR}
\def\confYear{2023}

\begin{document}

\title{Fix the Noise: Disentangling Source Feature for \\ Controllable Domain Translation}

\author{Dongyeun Lee$^{1,2}$ \quad Jae Young Lee$^{1}$ \quad Doyeon Kim$^{1}$ \quad Jaehyun Choi$^{1}$ \\\quad Jaejun Yoo$^{3}$ \quad Junmo Kim$^{1}$ \\ 
$^{1}$KAIST \qquad $^{2}$Klleon AI Research \qquad $^{3}$UNIST
}

\twocolumn[{%
\vspace{-2mm}
\renewcommand\twocolumn[1][]{#1}%
\maketitle
\vspace{-7mm}
\begin{center}
    \centering
    \begin{tabular}{c}
        \begin{minipage}{0.95\textwidth}
        \includegraphics[width=\linewidth]{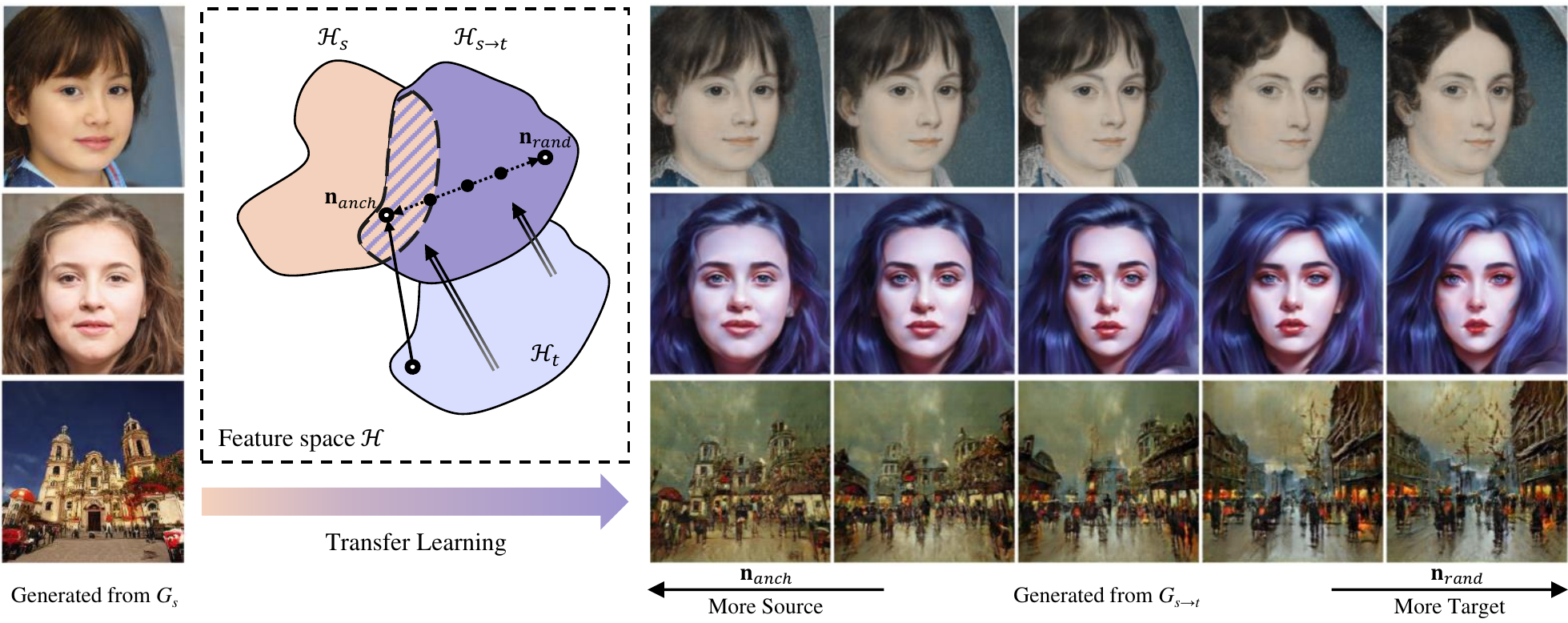}
        \end{minipage}
\hspace{-0.4cm}
\vspace{-0.1cm}
\\

\end{tabular}
\vspace{-0.2cm}
\captionof{figure}{Given a source model $G_s$, our method can smoothly control the degree of source domain features in a fine-tuned model $G_{s \rightarrow t}$. 
Samples in each row are generated from the same latent code $\mathbf z \in \mathcal Z$ by $G_s$ and $G_{s \rightarrow t}$. 
Here, $\mathcal H_s$, $\mathcal H_{s \rightarrow t}$, and $\mathcal H_t$ denote feature spaces of the source, our model, and simply fine-tuned target model, respectively. 
Our approach explicitly guides the model to preserve the source features by using the anchor point $n_{anch}$, which allows a flexible and smooth cross-domain control via $G_{s \rightarrow t}$.
}
\vspace{-0.2cm}
\end{center}%
}]

\maketitle

\begin{abstract}
\vspace{-4mm}
Recent studies show strong generative performance in domain translation especially by using transfer learning techniques on the unconditional generator.
However, the control between different domain features using a single model is still challenging.
Existing methods often require additional models, which is computationally demanding and leads to unsatisfactory visual quality.
In addition, they have restricted control steps, which prevents a smooth transition.
In this paper, we propose a new approach for high-quality domain translation with better controllability.
The key idea is to preserve source features within a disentangled subspace of a target feature space.
This allows our method to smoothly control the degree to which it preserves source features while generating images from an entirely new domain using only a single model. 
Our extensive experiments show that the proposed method can produce more consistent and realistic images than previous works and maintain precise controllability over different levels of transformation.
The code is available at \href{https://github.com/LeeDongYeun/FixNoise}{LeeDongYeun/FixNoise}.
\end{abstract}

\section{Introduction}
\label{sec:intro}
Image translation between different domains is a long-standing problem in computer vision \cite{gatys2016image,johnson2016perceptual,isola2017image,liu2017unsupervised,zhu2017unpaired,wang2018high,huang2018multimodal,choi2018stargan,choi2020stargan}.
Controllability in domain translation is important since it allows the users to set the desired properties. 
Recently, several studies have shown promising results in domain translation using a pre-trained unconditional generator, such as StyleGAN2 \cite{karras2020analyzing}, and its fine-tuned version \cite{pinkney2020resolution,lee2020freezeg,kwong2021unsupervised,song2021agilegan}.
These studies implemented domain translation by embedding an image from the source domain to the latent space of the source model and by providing the obtained latent code into the target model to generate a target domain image.
To preserve semantic correspondence between different domains, previous works commonly focused on the hierarchical design of the unconditional generator.
They used several techniques like freezing \cite{lee2020freezeg} and swapping \cite{pinkney2020resolution} layers or both \cite{kwong2021unsupervised}.
In these approaches, users can control the degree of preserved source features by setting the number of freezing or swapping layers of the target model differently.

However, one of the notable limitations of the previous methods is that they cannot control features across domains in a single model.
Imagine morphing between two images $x_0$ and $x_1$.
Previous methods approximated midpoints between $x_0$ and $x_1$ by either building a new hybrid model by converting weights or training a new model.
In these approaches, each intermediate point is drawn from the output distribution of different models, which would produce inconsistent results.
Moreover, getting an additional model for each intermediate point (image) also increases the computational cost.
Another common limitation of these layer-based methods is that their control levels are discrete and restricted to the number of layers, which prevents fine-grain control. 

In this paper, we introduce a new training strategy, FixNoise, for cross-domain controllable domain translation.
To control features across domains in a single model, we argue that the source features should be preserved but disentangled with the target in the model's inherited space.
To this end, we focus on the fact that the noise input of StyleGAN2, which is added after each convolution, expands the functional space composed of the latent code expression.
In other words, the feature space could be seen as a set of subspaces corresponding to each random noise.
To preserve the source features only to a particular subset of the feature space of the target model, we fix the noise input when applying a simple feature matching loss.
The disentangled feature space allows our method to fine-grain control the preserved source features only in a single model without limited control steps through linear interpolation between the fixed and random noise.
The extensive experiments demonstrate that our approach can generate more consistent and realistic results than existing methods on cross-domain feature control and also show better performance on domain translation qualitatively and quantitatively.

\section{Related work}
\noindent\label{sec:related work}
\textbf{Domain translation} aims to synthesize a target domain image conditioned on a source domain image.
Early works \cite{isola2017image,liu2017unsupervised,zhu2017unpaired,wang2018high} successfully solved domain translation by jointly training the encoder for the source and the decoder for the target.
Afterward, several works have extended this framework to multi-domain and multi-modal settings \cite{zhu2017toward,huang2018multimodal,lee2018diverse,liu2019few,choi2018stargan,choi2020stargan}.
On top of this framework, many studies have been conducted in diverse applications such as style transfer \cite{huang2017arbitrary,chen2017stylebank,shen2018neural,sanakoyeu2018style,Sheng_2018_CVPR}, cartoonization \cite{chen2018cartoongan,wang2020learning,kim2019u,nizan2020breaking,li2021anigan,su2021mangagan}, caricature generation \cite{li2020carigan,shi2019warpgan,gong2020autotoon}, and makeup transfer \cite{chang2018pairedcyclegan,gu2019ladn,jiang2020psgan,deng2021spatially}.
However, the joint training framework has a weakness in terms of scalability.
Since they train the network according to the source/target setting initially given, the entire framework has to be newly trained if it becomes a different setting, such as adding a new target domain.

\noindent\textbf{Domain translation using unconditional GANs.} Recently, several methods \cite{pinkney2020resolution,lee2020freezeg,kwong2021unsupervised,song2021agilegan} have introduced a new approach to domain translation by leveraging a pre-trained unconditional generator, such as StyleGAN2, of the source domain and that of the target domain which is fine-tuned from the source generator.
The new framework consists of a two-step approach for domain translation.
First, a latent code is obtained by embedding a source domain image to the latent space of the source generator by optimization \cite{ma2018invertibility,creswell2018inverting,Abdal_2019_ICCV,abdal2020image2stylegan++,karras2020analyzing} or encoder \cite{zhu2020indomain,xu2021generative,richardson2021encoding,10.1145/3450626.3459838,alaluf2021hyperstyle,dinh2021hyperinverter,wang2021HFGI,hu2022style}.
Then, a target domain image is generated by forwarding the given latent code to the target generator.
The success of this two-step approach is further explained by the observation of StyleAlign \cite{wu2022stylealign} that $\mathcal W$ space of the two models is similar, which is in line with the assumption of several methods in the joint training framework mentioned above \cite{liu2017unsupervised,huang2018multimodal,liu2019few}.
The two-step framework only requires different domain generators and the latent inversion method for domain translation.
It indicates that there is no need to train the entire framework for different settings.
Thus, the framework that utilizes the pre-trained unconditional generator is stable and superior to the joint training framework in terms of scalability.

In the two-step approach, previous methods introduced several techniques to encourage correspondence between images from different domains.
Layer-swap \cite{pinkney2020resolution} generated a target domain image with coarse spatial characteristics of the source domain by combining low-resolution layers of the source model and high-resolution layers of the target model.
By adjusting the number of layers to be swapped, their method can control the degree of remaining source features.
Freeze G \cite{lee2020freezeg} obtained a similar effect by freezing weights of initial layers of the generator during transfer learning.
UI2I StyleGAN2 \cite{kwong2021unsupervised} froze mapping layers to ensure exactly the same $\mathcal{W}$ space between the source and target models and combined it with Layer-swap. 
AgileGAN \cite{song2021agilegan} tried to preserve the source domain features by early stopping.
Some recent studies \cite{yang2022pastiche,yang2022Vtoonify} introduce an exemplar-based task in a limited data setup.
However, the exemplar-based task requires latent optimization that is substantially time-consuming \cite{richardson2021encoding} for the entire dataset, which is not practically applicable to large datasets.

\section{Method}
\label{sec:method}
In StyleGAN2, the model synthesizes an output image $x$ from a latent code $\mathbf z \in \mathcal Z$ and a noise input $\mathbf n \sim \mathcal N(\mathbf 0, \mathbf I)$, which is expressed as
\begin{equation}\label{eq:1}
    x = G(\mathbf z, \mathbf n).
\end{equation}
Given a source domain model $G_s$, our goal is to train a target domain model $G_{s \rightarrow t}$ initialized with the source model weights while preserving the source domain features.
In Sec.~\ref{sec:loss}, we briefly discuss the space in which to preserve features and introduce a simple but effective feature matching loss.
In Sec.~\ref{sec:fixnoise}, we propose FixNoise that ensures disentanglement between the two domain features in the feature space of the target model.
In Sec.~\ref{sec:interpolation}, we introduce cross-domain feature control using noise interpolation.

\subsection{Which feature to preserve?}
\label{sec:loss}
Remark that StyleGAN2 \cite{karras2020analyzing} contains two types of feature spaces: an intermediate feature space that consists of feature convolution layer outputs and an RGB space that consists of RGB outputs transformed from an intermediate feature by tRGB layers.
We choose to preserve the intermediate feature space, which is denoted as $\mathcal H$, for the following reasons.
First, it has recently been found that the feature convolution layers
change the most among layers during transfer learning \cite{wu2022stylealign}.
This observation indicates that the source features mostly vanish in $\mathcal H$ which is the output space of the feature convolution layers.
Second, matching features of the source and target models in $\mathcal H$ enables the subsequent tRGB layers to learn the target distribution.
Consequently, preserving $\mathcal H$ space when training enables $G_{s \rightarrow t}$ to maintain coarse features of the source while learning fine features of the target. 
Furthermore, images generated from the model trained with such preservation go beyond simple color filtering effects applied on source images.

From the same latent code $\mathbf z \in \mathcal Z$, we encourage the target model $G_{s \rightarrow t}$ to have similar features as those of the source model $G_s$ in $\mathcal H$ using a simple feature matching loss
\begin{equation}\label{eq:fm}
    \mathcal L_{fm} = \mathbb E_{\mathbf z} \Bigg[ \frac{1}{L}\sum_{l=0}^L \big(G_s^l(\mathbf z, \mathbf n_s) -G_{s \rightarrow t}^l(\mathbf z, \mathbf n_{s \rightarrow t})\big)^2 \Bigg],
\end{equation}
where $L$ denotes the number of feature convolution layers. Note that $\mathbf n_s$ and $\mathbf n_{s \rightarrow t}$ are independently sampled noise inputs for each model, respectively.
Recall that losses that utilize the intermediate features of a network are widely used in GANs literature, such as perceptual loss \cite{johnson2016perceptual,zhang2018perceptual}.
However, the main difference between $\mathcal L_{fm}$ and the perceptual loss is in which space the features are matched.
The perceptual loss 
encourages the source and target models to have similar features in the feature space of the external network which is unrelated to image generation, whereas our loss encourages them to have similar intermediate features internally.
With the loss $\mathcal L_{fm}$, we can encourage the target model to have a shared feature space with the source model internally.

\begin{figure}[t]
    \centering
    \begin{tabular}{ccc}
    \begin{minipage}{0.19\textwidth}\includegraphics[trim=0cm 0cm 0cm 0cm,clip,width=\linewidth]{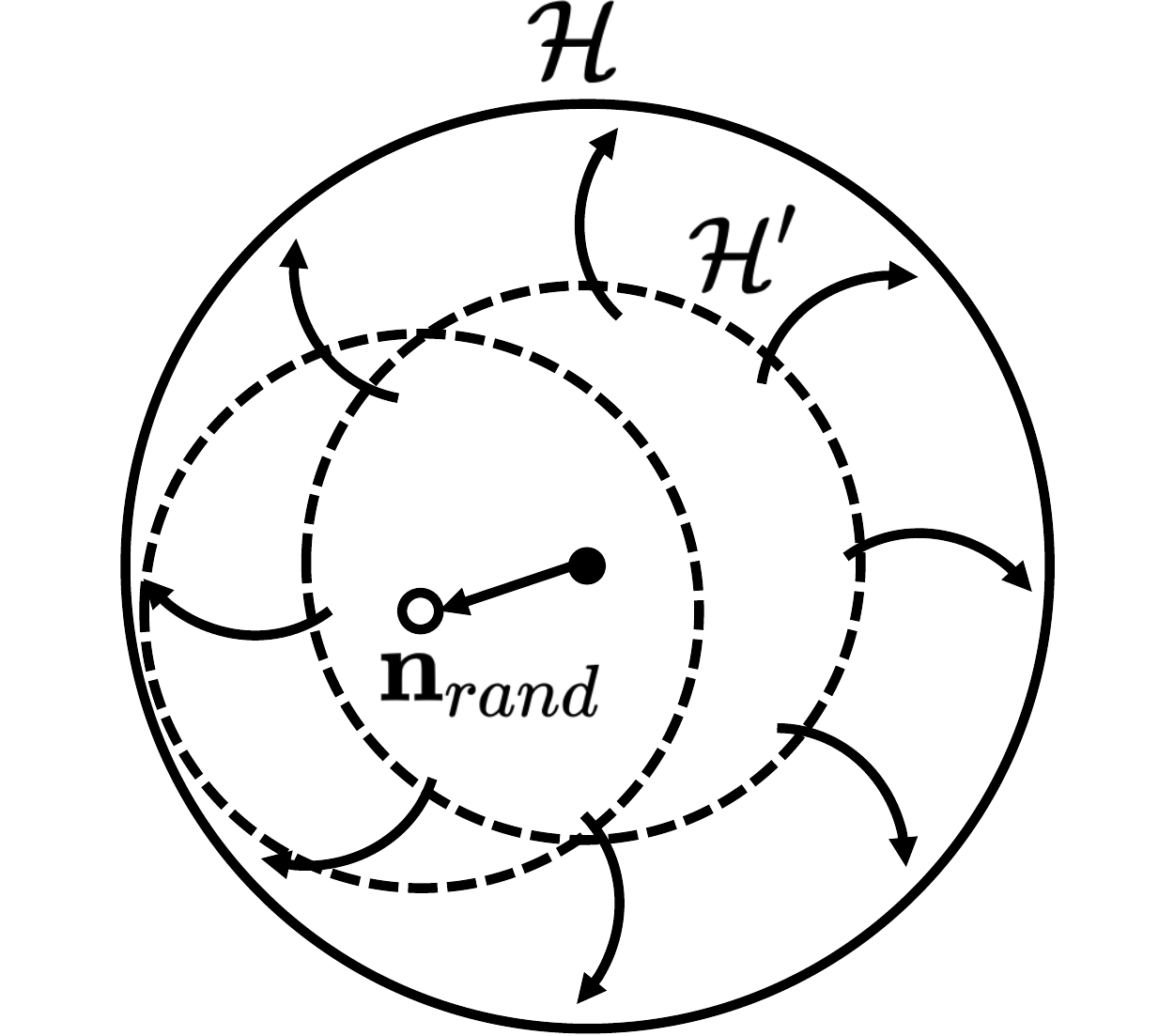}
    \end{minipage} &
    \begin{minipage}{0.19\textwidth}\includegraphics[trim=0cm 0cm 0cm 0cm,clip,width=\linewidth]{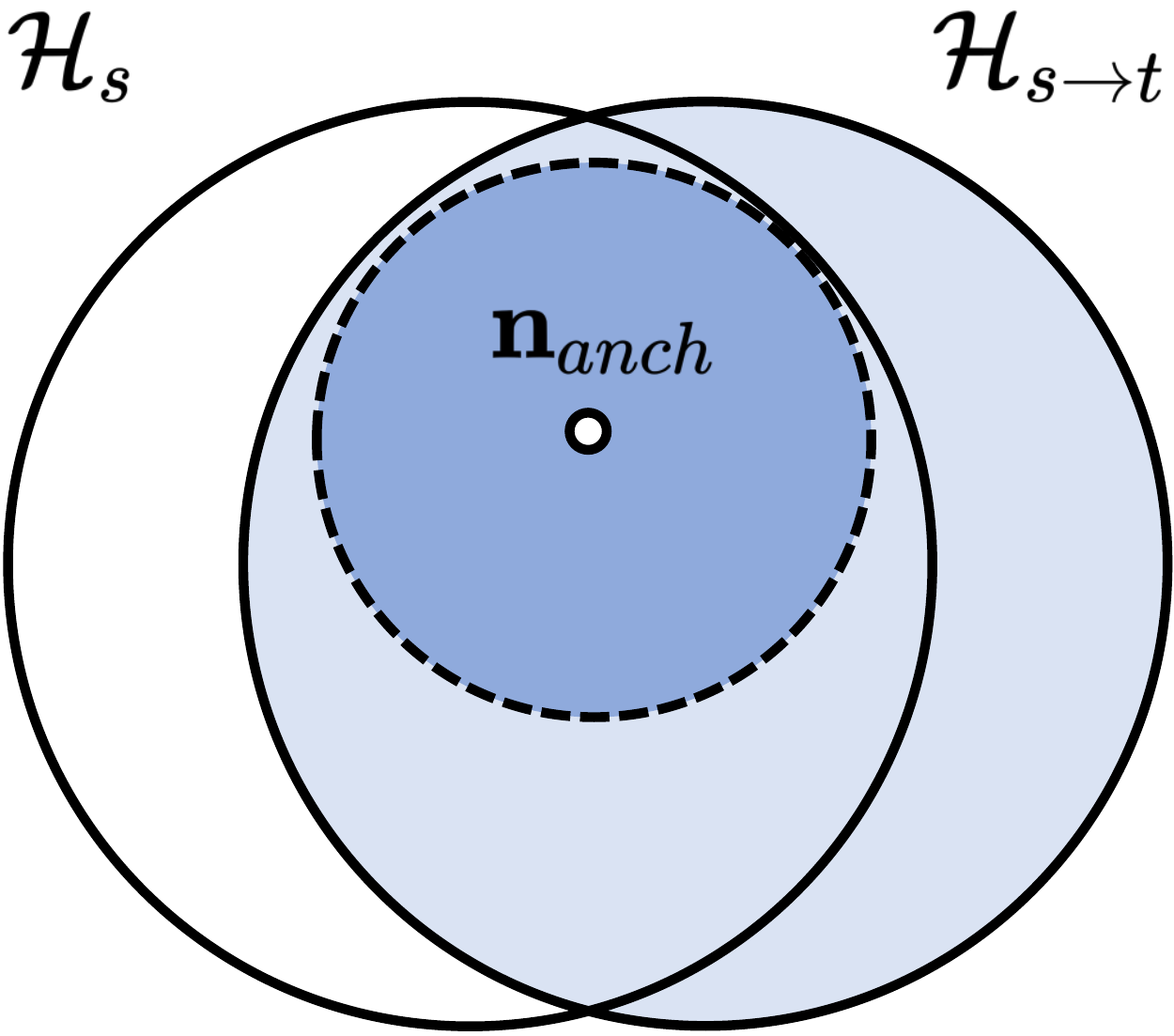}
    \end{minipage}
    \vspace{3mm}
\\
 
(a) & 
(b)  \\

\end{tabular}
\caption{An illustration of FixNoise. 
(a) The black dot indicates $\mathbf 0$ noise corresponding to $\mathcal H'$. 
 Randomly sampled noise expands $\mathcal H'$ to $\mathcal H$.
(b) Anchored subspace is denoted by a dotted line. Source features are only mapped to the anchored subspace of $\mathcal H_{s \rightarrow t}$.}
\label{fig:2}
\end{figure}

\subsection{Disentangled feature space using FixNoise}
\label{sec:fixnoise}
The loss $\mathcal L_{fm}$ enforces the entire feature space of the target model $\mathcal H_{s \rightarrow t}$ to be the same as that of the source model when we naively apply the loss.
This may disturb $G_{s \rightarrow t}$ to learn diverse target features that do not exist in the source domain.
Even if the target features are learned, the degree of preserved source features cannot be controlled if the source and target features are entangled in the feature space of the target model.
Instead of applying the loss $\mathcal L_{fm}$ to the entire feature space $\mathcal H_{s \rightarrow t}$, we introduce an effective strategy, FixNoise, that does not disturb target feature learning and allows the different domain features to be disentangled from each other in the feature space of the target model.
Our method begins with an assumption that both can be achieved if the source features are mapped in a particular subspace in $\mathcal H_{s \rightarrow t}$.

\begin{figure*}[t!]
\centering
\includegraphics[width=1\linewidth]{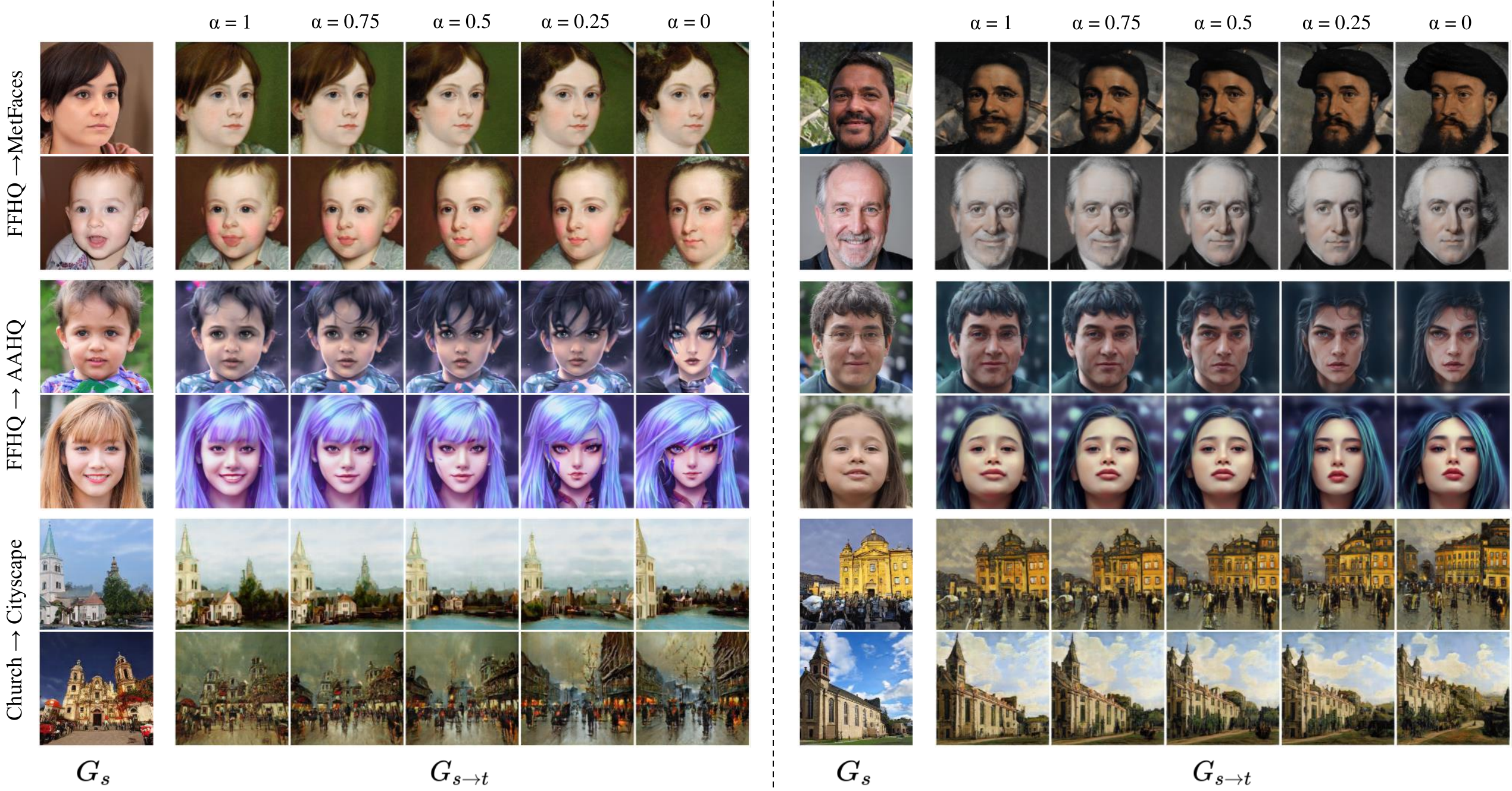}
  \caption{Noise interpolation results on different settings. 
  The interpolation weight $\alpha$ is presented above each column.}
\label{fig:blending}
\end{figure*}

As Eq.~\ref{eq:1}, a final output image is generated from two input components: the latent code $\mathbf z \in \mathcal Z$, and the per-pixel Gaussian noise $\mathbf n \sim \mathcal N(\mathbf 0, \mathbf I)$ which creates stochastic variation such as curls of hair, eye reflection, and background detail.
If the noise $\mathbf n = \mathbf 0$, a feature $\mathbf h' \in \mathcal H'$ is deterministically generated from a latent code $\mathbf z$, where $\mathcal H'$ is a subspace of $\mathcal H$ corresponding to $\mathbf n= \mathbf 0$.
As depicted in Figure~\ref{fig:2}, when each randomly sampled noise $\mathbf n_{rand}$ is added to $\mathbf h'$, it shifts $\mathcal H'$ to a space that corresponds to each $\mathbf n_{rand}$.
This shift of subspace by each noise input consequently expands $\mathcal H'$ to $\mathcal H$.
It signifies that the feature space $\mathcal H$ consists of subspaces corresponding to each random noise $\mathbf n_{rand}$.
To ensure that the source features are only mapped to a particular subspace of $\mathcal H_{s \rightarrow t}$, we fix the noise to a single predefined value when 
$\mathcal L_{fm}$ is applied.
By substituting $\mathbf n_s$ and $\mathbf n_{s \rightarrow t}$ to $\mathbf n_{anch}$ in Eq.~\ref{eq:fm}, the feature matching loss $\mathcal L_{fm}$ with FixNoise is described as
\begin{equation} \label{eq:fm_fixnoise}
    \mathcal L_{fm}' = \mathbb E_{\mathbf z} \Bigg[ \frac{1}{L}\sum_{l=0}^L \big(G_s^l(\mathbf z, \mathbf n_{anch}) -G_{s \rightarrow t}^l(\mathbf z, \mathbf n_{anch})\big)^2 \Bigg],
\end{equation}
where $\mathbf n_{anch}$ denotes the fixed noise.
We refer to the fixed noise as \textit{anchor point} $\mathbf n_{anch}$ and the corresponding subspace as \textit{anchored subspace}.
$\mathbf n_{anch}$ is sampled from the Gaussian distribution same as $\mathbf n_{rand}$ and fixed for the whole training process.
The anchor point $\mathbf n_{anch}$ gives the model explicit guidance to preserve the source feature in $\mathcal H_{s \rightarrow t}$.

To learn the target features over the entire feature space $\mathcal H_{s \rightarrow t}$, we use randomly sampled noise $\mathbf n_{rand}$ when applying the adversarial loss \cite{goodfellow2014generative}
\begin{equation} \label{eq:adv}
    \mathcal L_{adv} = \mathbb E_{\mathbf z} \big[ - \log D(G_{s \rightarrow t}(\mathbf z, \mathbf n_{rand})) \big],
\end{equation}
where $D$ denotes a discriminator.

By combining Eq.~\ref{eq:fm_fixnoise} and Eq.~\ref{eq:adv}, our objective function for $G_{s \rightarrow t}$ is described as
\begin{equation}
    \mathcal L_{total} = \mathcal L_{adv} + \lambda \mathcal L_{fm}',
\end{equation}
where $\lambda$ denotes loss weight.
Through this, the source features are only mapped in the anchored subspace by  $\mathcal L_{fm}'$, while the target features are freely adapted to the entire $\mathcal H_{s \rightarrow t}$ by $\mathcal L_{adv}$.
We believe that common features of the two domains are embedded in the anchored subspace, and features that exist only in the target are embedded in the remainder space of $\mathcal H_{s \rightarrow t}$.
To sum up, the disentanglement between the different domain features can be achieved in $\mathcal H_{s \rightarrow t}$ by the anchor point.

\subsection{Cross-domain feature control}
\label{sec:interpolation}
As described in Sec.~\ref{sec:fixnoise}, we achieved disentanglement between the two domains within the feature space of the target model.
To be specific, we preserve the source features only in the anchored subspace of $\mathcal H_{s \rightarrow t}$ that corresponds to the fixed noise $\mathbf n_{anch}$.
On the other hand, the target features are learned to the entire $\mathcal H_{s \rightarrow t}$ that corresponds to all random noise $\mathbf n_{rand}$.
What should be noted here is that only noise input disentangles the two domain features in $\mathcal H_{s \rightarrow t}$.
This enables a smooth transition between images by linear interpolation of the anchor point and random noise:
\begin{equation}
    \mathbf n_{interp} = \alpha \cdot \mathbf n_{anch} + (1-\alpha) \cdot \mathbf n_{rand},
\end{equation}
\vspace{-2mm}
\begin{equation}
    x_{interp} = G_{s\rightarrow t}(\mathbf z, \mathbf n_{interp}),
\end{equation}
where $\mathbf n_{interp}$ is interpolated noise and $\alpha$ represents the interpolation weight.
This property is in line with recent work \cite{Liu_2021_CVPR} that enables smooth transition across domains by interpolation of latent code.
The main assumptions of their work are that the smooth transition by the latent interpolation can be achieved if (i) the margin between the different domains in latent space is small, and (ii) the entire latent distribution is Gaussian.
The fact that the fixed and random noise are sampled from Gaussian distribution already satisfies their two assumptions.
Thus, our approach, which utilizes Gaussian noise to disentangle different domain features, enables a gradual transition between the two domains as shown in Figure~\ref{fig:blending}.

\begin{figure}[t!]
\centering
\includegraphics[width=1\linewidth]{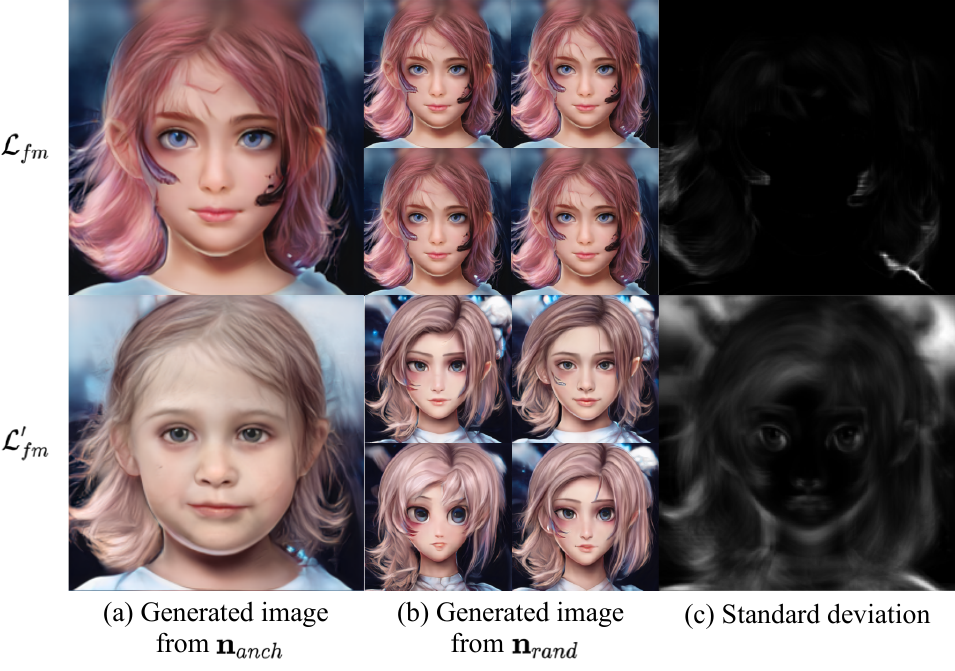}
  \caption{Visualizing the effect of the noise input when FixNoise is applied. 
  Images in the same training setting are generated from the same latent code $\mathbf z$. 
  The first and second row correspond to results from $G_t$ trained by applying $\mathcal L_{fm}$ (Eq.~\ref{eq:fm}) and $\mathcal L_{fm}'$ (Eq.~\ref{eq:fm_fixnoise}), respectively. 
  (a) Generated images from the anchor point $\mathbf n_{anch}$. (b) Generated images from the random noise $\mathbf n_{rand}$.
  Varying the noise has global effects such as identity or structure when FixNoise is applied (zoom-in is recommended).
  (c) Standard deviation of each pixel over 100 different random noise inputs.
  The FixNoise strategy makes the noise affect images more coarsely.}
\label{fig:noisetype}
\vspace{-3mm}
\end{figure}

\section{Experiments}

\begin{figure*}[t!]
\centering
\includegraphics[width=1\linewidth]{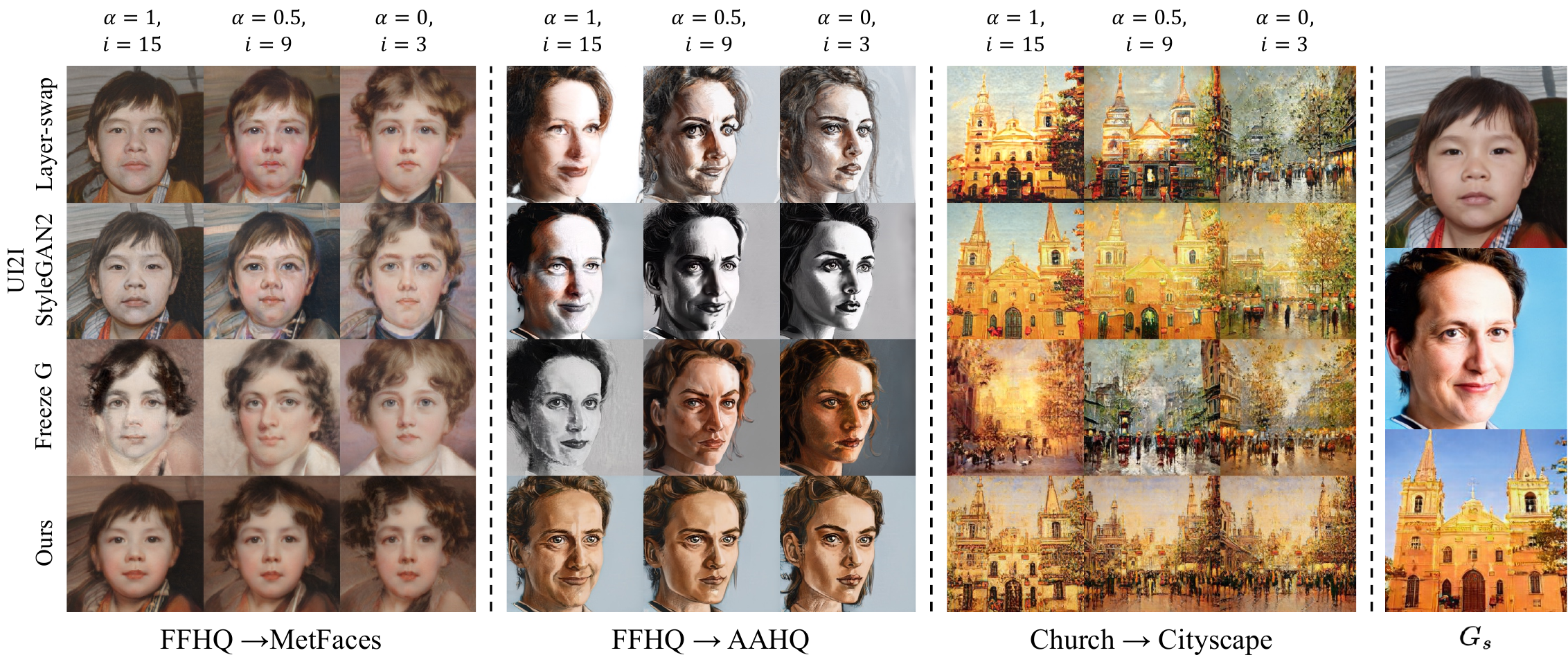}
  \caption{Qualitative comparison on controlling  preserved source features.
  Our results show a consistent transition between the source and target features.}
\label{fig:compare_baseline}
\end{figure*}

\noindent\textbf{Datasets.}
We conduct experiments for several source and target settings considering the spatial similarity between the source and target domains.
For the similar domain setting, we transfer FFHQ \cite{karras2019style} to MetFaces \cite{Karras2020ada} and AAHQ \cite{liu2021blendgan}.
For the distant domain setting, we transfer LSUN Church \cite{yu2015lsun} to WikiArt Cityscape \cite{wikiart}.
All experiments are conducted on $256 \times 256$ resolution images.

\noindent\textbf{Implemention detail.}
We build upon the base configuration in the official Pytorch \cite{paszke2019pytorch} implementation of StyleGAN2-ADA\footnote{\url{https://github.com/NVlabs/stylegan2-ada-pytorch}} \cite{Karras2020ada}.
Official pre-trained weights for the source model and discriminator trained on FFHQ\footnote{\url{http://nvlabs-fi-cdn.nvidia.com/stylegan2-ada-pytorch/pretrained/transfer-learning-source-nets/ffhq-res256-mirror-paper256-noaug.pkl}} and LSUN Church\footnote{\url{http://nvlabs-fi-cdn.nvidia.com/stylegan2/networks/stylegan2-church-config-f.pkl}} are used.
We use a non-saturating adversarial loss \cite{goodfellow2014generative} for $\mathcal L_{adv}$ and set $\lambda = 0.05$ for all experiments.
Following StyleGAN2 \cite{karras2020analyzing}, we additionally use style mixing regularization \cite{karras2019style} and path length regularization \cite{karras2020analyzing} with $\mathcal L_{total}$. 
The discriminator objective function follows StyleGAN2-ADA.
Like $G_{s \rightarrow t}$, the discriminator $D$ is also initialized with the weights of the source model's discriminator.
Adaptive discriminator augmentation \cite{Karras2020ada} is used to prevent discriminator overfitting.
The batch size is set to 64.
FFHQ $\rightarrow$ MetFaces, FFHQ $\rightarrow$ AAHQ, and LSUN Church $\rightarrow$ WikiArt Cityscape are trained for 2000K, 12000K, and 5000K images, respectively.

\setlength{\tabcolsep}{5.5pt}
\begin{table}[t!]
\begin{center}
\small
\begin{tabular}{c|cc|cc|cc}
\toprule
Source        & \multicolumn{4}{c|}{FFHQ}                                  & \multicolumn{2}{c}{Church}                \\ \midrule
Target        & \multicolumn{2}{c|}{MetFaces}  & \multicolumn{2}{c|}{AAHQ} & \multicolumn{2}{c}{Cityscape}             \\ \midrule
$\alpha$& LPIPS          & FID              & LPIPS         & FID               & LPIPS             & FID             \\ \midrule
1       & \textbf{0.412} & 40.37            & \textbf{0.316}& 31.65             & \textbf{0.521}    & 27.64    \\
0.75    & 0.432          & 37.59            & 0.366         & 22.70             & 0.557             & 20.59             \\
0.5     & 0.451          & 30.17            & 0.381         & 14.60             & 0.626             & 17.37             \\
0.25    & 0.481          & 23.27            & 0.410         & 13.65             & 0.653             & 12.53             \\
0       & 0.536          & \textbf{19.68}   & 0.510         & \textbf{5.10}     & 0.679             & \textbf{11.49}             \\
\bottomrule
\end{tabular}
\caption{Quantitative comparison with different interpolation weights. We report the best FID, and measure LPIPS using the same network snapshot.
}
\label{table:noise_blend}
\end{center}
\vspace{-5mm}
\end{table}
\setlength{\tabcolsep}{0.5pt}

\setlength{\tabcolsep}{10pt}
\begin{table*}[ht]
\begin{center}
\small
\begin{tabular}{c|crr|crr|crr}
\toprule
Source        & \multicolumn{6}{c|}{FFHQ}                                  & \multicolumn{3}{c}{Church}        \\ \midrule
Target        & \multicolumn{3}{c|}{MetFaces}  & \multicolumn{3}{c|}{AAHQ} & \multicolumn{3}{c}{Cityscape}     \\ \midrule
  &  PS& FID & \vtop{\hbox{KID} \hbox{\tiny{($\times 10^3)$}}} & PS & FID & \vtop{\hbox{KID} \hbox{\tiny{($\times 10^3)$}}} & PS & FID & \vtop{\hbox{KID} \hbox{\tiny{($\times 10^3)$}}} \\
\midrule
Layer-swap              & 0.641 & 68.31 & 34.69                & 0.574 &38.03 & 28.30              & 0.604 &52.02 & 38.46                 \\ 
UI2I StyleGAN2          & 0.649 & 79.54 & 45.38                & 0.594 &51.10 & 40.77              & 0.596 &64.49 & 50.03                \\ 
Freeze G                & 0.496 & 24.12 & 5.41                 & 0.465 &7.93 &  2.82              & 0.446 &12.84 & 3.55                 \\ 
Ours ($\alpha = 1$)     & \multirow{2}{*}{\textbf{0.828}} & 40.37 & 14.88               & \multirow{2}{*}{\textbf{0.835}} & 31.65 & 22.86   & \multirow{2}{*}{\textbf{0.709}}  & 27.64 & 16.28                 \\
Ours ($\alpha = 0$)     &  &\textbf{19.68} & \textbf{3.31}   &  & \textbf{5.10} & \textbf{1.55} &  & \textbf{11.49} & \textbf{3.03}  \\
\midrule\midrule
StyleGAN2-ADA        &  \xmark  & 19.04 & 2.74 &  \xmark  &    4.32 & 1.22 &  \xmark  &  11.04 & \ \,2.75  \\
\bottomrule
\end{tabular}
\caption{Quantitative comparison with unconditional GANs based methods. We report the best FID, and measure PS and KID using the same network snapshot.}
\label{table:2}
\end{center}
\vspace{-5mm}
\end{table*}
\setlength{\tabcolsep}{0.5pt}

\subsection{Analysis of FixNoise}
\noindent\textbf{Effect of the noise input.}
In the original StyleGAN model \cite{karras2019style,karras2020analyzing}, the latent code affects global aspects such as identity and pose, whereas the noise input affects inconsequential stochastic variation (\textit{e.g.} curls of hair, eye reflection, and background detail) which is a localized effect.
However, when we apply FixNoise during transfer learning of the model, we find that the noise gives more diverse effects to the images.
Figure~\ref{fig:noisetype} shows how FixNoise changes the effect of the noise input on the generated images.
We can observe that the noise input also affects global aspects when FixNoise is applied, whereas, without FixNoise, the noise only affects stochastic aspects.
The observation is due to the discrepancy between the source and target domains.
The discrepancy between different domain features includes not only the local but also global aspects.
As mentioned above, the source and target features are disentangled in the feature space $\mathcal H_{s \rightarrow t}$, and the only factor that disentangles the two domain features is the noise input.
Thus, the noise input, which is responsible for the disentanglement between two different domain features, is given the role to control some global aspects.

\noindent\textbf{Noise interpolation.}
We evaluate our effectiveness on cross-domain feature control in different training settings.
As shown in Figure~\ref{fig:blending}, the source domain features are well preserved in the images generated from the anchored subspace ($\alpha=1$), whereas they are lost in the rest of space in $\mathcal H_{s \rightarrow t}$ ($\alpha=0$).
This indicates that FixNoise successfully disentangles the source and target features in $\mathcal H_{s \rightarrow t}$.
The fact that the features of both domains are embedded in a single space $\mathcal H_{s \rightarrow t}$ enables a smooth transition between the source and target features through interpolation between the anchor point $\mathbf n_{anch}$ and other randomly sampled noise $\mathbf n_{rand}$.
This allows us to control the degree of preserved source features.
Further, we quantitatively examine the effects of the noise interpolation. 
LPIPS \cite{zhang2018perceptual} and FID \cite{heusel2017gans} are used to capture distance with the source and target distribution, respectively.
For LPIPS, we randomly sample 2000 latent codes $\mathbf z \in \mathcal Z$ and measure the distance between images generated by $G_s$ and $G_{s \rightarrow t}$ from the same $\mathbf z$.
We use target domain images for FID measurement.
Low LPIPS indicates that the generated images are similar to the source images, while low FID indicates that the distribution of the generated images is close to the target data distribution.
As shown in Table~\ref{table:noise_blend}, our method obtained the lowest FID and highest LPIPS when $\alpha = 0$, and vice versa when $\alpha = 1$.
Thus, we could infer that the generated images lose their source 
features and approach the target distribution as $\alpha$ decreases.
The result demonstrates that our method can control features across domains just by modifying the noise weight.

\begin{figure*}[t!]
\centering
\includegraphics[width=1\linewidth]{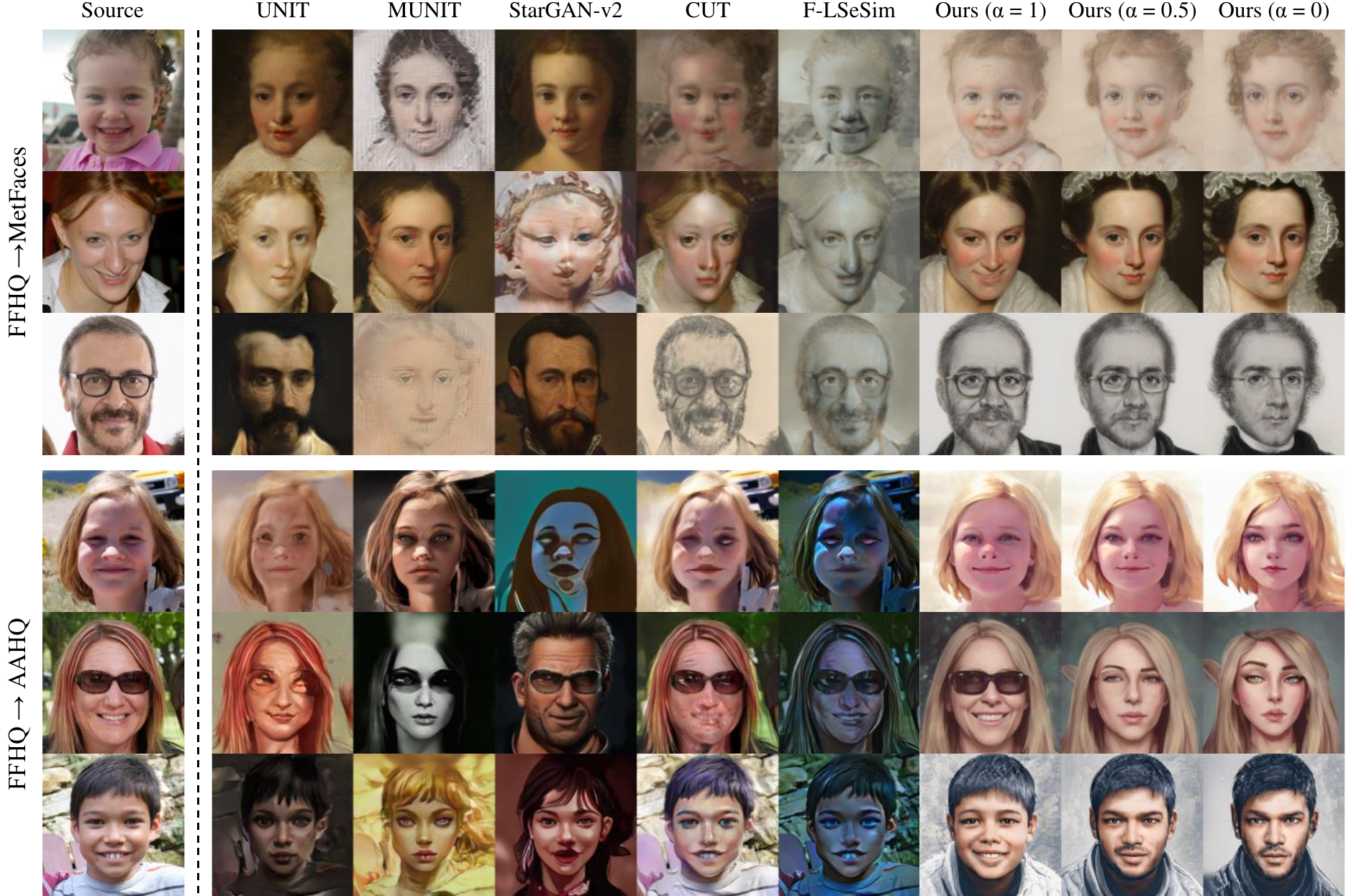}
  \caption{Qualitative comparison with domain translation methods.}
\label{fig:compare_DT}
\end{figure*}

\subsection{Comparison}
We evaluate the proposed method with two different approaches: unconditional GANs based methods and conventional domain translation methods. The detailed evaluation metrics are described in the supplemental material.

\noindent\textbf{Comparison with unconditional GANs based method.}
We compare our approach with unconditional GANs based approaches for domain translation including Freeze G \cite{lee2020freezeg}, Layer-swap \cite{pinkney2020resolution} and UI2I StyleGAN2 \cite{kwong2021unsupervised} that combines freeze FC with Layer-swap.
These methods including ours have constraints on source feature preservation.
In order to explore how the constraints of each method affect target distribution learning, we additionally train StyleGAN2-ADA \cite{Karras2020ada} under the same source and target settings.
Note that StyleGAN2 has 21 layers including constant input for $256 \times 256$ resolution images.
When freezing or swapping layer $i = 0$, $G_{s \rightarrow t}$ is fine-tuned without any constraints, and when $i = 21$, $G_{s \rightarrow t}$ is mere $G_s$.

The qualitative comparison with previous leading methods is shown in Figure~\ref{fig:compare_baseline}.
For a qualitative comparison, we use the interpolation weight $\alpha = 1, 0.5, 0$ for ours, and $i = 15, 9, 3$ for baselines to match a similar preservation level, respectively.
In  FFHQ $\rightarrow$ MetFaces and FFHQ $\rightarrow$ AAHQ settings,
inconsistent transitions occur in the baselines.
In particular, changes in human identity are observable in the results of Layer-swap. 
Several unnatural color transitions are observed in Layer-swap and UI2I StyleGAN2 due to simply combining two different models.
Although the inconsistency and color transition problems are less important in Church $\rightarrow$ Cityscape, the feature control steps in the baselines are restricted to the number of layers, which interferes with a fine-grain transition.
In addition, to control the source features, previous methods require new models by swapping layers or training, which is not suitable for practical application.
In contrast, our method that generates the most realistic results enables smooth transition in a single model, which is easily applicable to diverse tasks.

The quantitative comparison is shown in Table~\ref{table:2}.
We use a modified version of Perceptual Smoothness (PS) \cite{Liu_2021_CVPR} to measure the smoothness of interpolation between different domain features.
FID \cite{heusel2017gans} and KID \cite{binkowski2018demystifying} are adopted to evaluate generation quality and diversity.
For PS, we use the interpolation weight $\alpha = 1, 0.75, 0.5, 0.25, 0$ for ours, and $i = 15, 12, 9, 6, 3$ for baselines to get interpolated images from the same $\mathbf z \in \mathcal Z$, respectively.
For FID and KID, we use $i = 15$ for baselines following \cite{kwong2021unsupervised}.
The PS of our method notably outperforms the competing methods, which implies that our approach is more precise and consistent in controlling features across domains.
Moreover, when $\alpha = 0$, our method achieved the highest FID and KID.
Compared to StyleGAN2-ADA which does not have any constraints on source preservation, our method shows similar performance while the other methods show significant performance drops.
It indicates that, in contrast to previous methods, our proposed method hardly interferes with the learning of the target distribution.
Although Freeze G obtained a better FID and KID than ours when $\alpha = 1$, they require additional models for each control level and show inconsistent results as shown in Figure~\ref{fig:compare_baseline}.
In short, our approach can produce the most coherent and fine-grain interpolation results while generating the most realistic images.

\setlength{\tabcolsep}{10pt}
\begin{table}[t!]
  \centering
    \begin{tabular}{c|rr|rr}
\toprule
\small{Source}        & \multicolumn{4}{c}{\small{FFHQ}}                                        \\ \midrule
\small{Target}        & \multicolumn{2}{c|}{\small{MetFaces}}  & \multicolumn{2}{c}{\small{AAHQ}}  \\ \midrule
  &  \small{FID} & \vtop{\hbox{\small{KID}} \hbox{\tiny{($\times 10^3)$}}}  & \small{FID} & \vtop{\hbox{\small{KID}} \hbox{\tiny{($\times 10^3)$}}}  \\
\midrule
\small{UNIT}    &   \small{42.69} & \small{22.66}   &   \small{20.12} & \small{14.78}      \\
\small{MUNIT}    &   \small{93.77} & \small{81.73}   &   \small{21.82} & \small{16.73}       \\
\small{StarGAN-v2}         &   \small{37.61} & \small{17.88}    &    \small{19.20} &  \small{\textbf{10.44}}   \\
\small{CUT}         &   \small{55.52} & \small{34.04}    &    \small{20.29} &  \small{12.26}   \\
\small{F-LSeSim}         &   \small{71.07} & \small{47.74}    &    \small{47.10} &  \small{38.90}   \\
\small{Ours ($\alpha = 1$)}     &   \small{53.80} & \small{31.45}    &       \small{34.90} & \small{23.46}        \\
\small{Ours ($\alpha = 0$)}     &   \small{\textbf{27.14}} & \small{\textbf{10.29}}    &       \small{\textbf{18.53}} & \small{11.75}   \\
\bottomrule
\end{tabular}
  \caption{Quantitative comparison with domain translation methods.
  We report the best FID, and measure KID using the same network snapshot.}
  \label{table:compare_DT}
  \vspace{-3mm}
  \end{table}
\setlength{\tabcolsep}{0.5pt}

\noindent\textbf{Comparison with domain translation method.}
We additionally compare our method to recent domain translation methods including UNIT \cite{liu2017unsupervised}, MUNIT \cite{huang2018multimodal}, StarGAN-v2 \cite{choi2020stargan}, CUT \cite{park2020cut}, F-LSeSim \cite{zheng2021spatiallycorrelative} by combining ours with the inversion method.
It has recently been observed that an optimization method to $\mathcal Z$ space shows the best performance in domain translation among inversion methods to several spaces (\textit{e.g.} $\mathcal Z+, \mathcal W$, and $\mathcal W+$) \cite{wu2022stylealign}.
They observed that inversion to $\mathcal W$ or $\mathcal W+$ space yields good reconstruction, but causes color artifacts to target images due to the changes in mapping function ($\mathcal Z$ to $\mathcal W$) when training the target model.
Thus, following StyleAlign \cite{wu2022stylealign}, we modify the optimization method from StyleGAN2 \cite{karras2020analyzing} to embed source images into $\mathcal Z$ space of the source model.

Figure~\ref{fig:compare_DT} shows the qualitative comparison on the domain translation task.
Competing methods except for StarGAN-v2 commonly fail to generate realistic images, and particularly include remarkable artifacts in generated images.
Results of StarGAN-v2 show better visual quality than the other competing methods, however, some of them are unnatural and fail to preserve the human identity.
Although CUT and F-LSeSim successfully preserve the identity of the source images, they generate source images with simple filtering effects and did not adapt well to the target domain.
Compared to the competing methods, our approach generates the most realistic and well-adapted images while preserving the source features.
In addition to visual quality, the most notable property of our approach compared to previous methods is that the preserved source features can be controlled in only a single model.

The quantitative comparison is shown in Table~\ref{table:compare_DT}.
We randomly sample 20K images from the source domain and generate a single image from each image.
When $\alpha = 1$, our method got higher FID and KID than UNIT and StarGAN-v2.
However, since FID and KID measure the distance from the target distribution, it is natural that the FID and KID are high when the source features are strongly preserved.
We emphasize that the advantage of our method is that we can control the degree of the preserved source features.
When the source features are less preserved ($\alpha = 0$), our method greatly outperforms other competing methods except for KID of FFHQ $\rightarrow$ AAHQ. 
Also, compared to StarGAN-v2, which achieved the best performance among competing methods, our method is not only qualitatively good, but also preserves the features of the source images much better.

\begin{figure}[t!]
    \centering
    \includegraphics[width=1\linewidth]{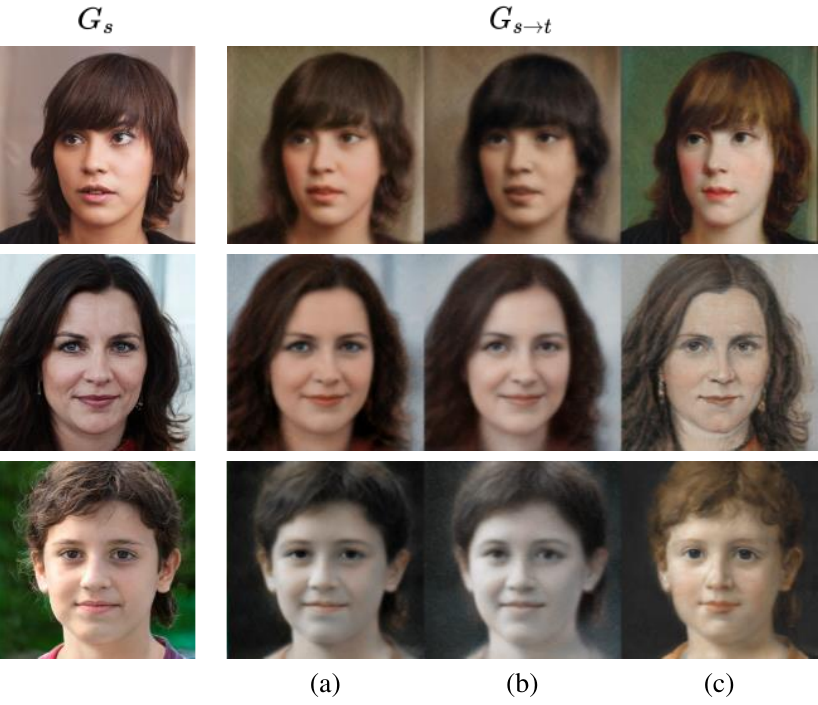}
    \caption{Visualizing the effects of the feature matching loss in different spaces: (a) image space, (b) RGB space, (c) intermediate feature space $\mathcal H$ (ours).}
    \label{fig:space}
  \vspace{-3mm}
\end{figure}

\subsection{Ablation study on feature matching loss}
\label{sec:ablation}
In FFHQ $\rightarrow$ MetFaces setting, we study in which space features are appropriate to preserve.
We conduct an experiment by applying a feature matching loss in intermediate feature space $\mathcal H$ (ours), RGB space, and image space.
Figure~\ref{fig:space} shows a qualitative comparison on applying the loss in the different spaces.
Results of the loss applied in the image and RGB space show well-preserved source features, however, they are not adapted to the target domain.
On the other hand, when the loss is applied to the intermediate feature space (ours), a target model $G_{s \rightarrow t}$ successfully learns features of the target domain while preserving the source features.
The feature matching loss in $\mathcal H$ allows tRGB layers to learn the target distribution.


\section{Conclusion}
In this paper, we proposed a new training strategy, FixNoise, 
for cross-domain controllable domain translation.
By focusing on the fact that the noise input of StyleGAN2 expands the functional space composed of the latent expression, our approach successfully disentangles the source and target features in the feature space of the target model.
Consequently, through the noise interpolation, our method can coherently control the degree of the source features in a single model without limited steps.
Furthermore, experimental results show that the proposed method remarkably outperforms the previous works in terms of image quality and consistent transition.
We believe that our methods can be applied to various fields that utilize multi-domain features.
Additional future work and limitations are described in the supplemental material.

{\small
\bibliographystyle{ieee_fullname}
\bibliography{egbib}

\begin{thebibliography}{10}\itemsep=-1pt

\bibitem{Abdal_2019_ICCV}
Rameen Abdal, Yipeng Qin, and Peter Wonka.
\newblock Image2stylegan: How to embed images into the stylegan latent space?
\newblock In {\em Proceedings of the IEEE/CVF International Conference on
  Computer Vision (ICCV)}, October 2019.

\bibitem{abdal2020image2stylegan++}
Rameen Abdal, Yipeng Qin, and Peter Wonka.
\newblock Image2stylegan++: How to edit the embedded images?
\newblock In {\em Proceedings of the IEEE/CVF Conference on Computer Vision and
  Pattern Recognition}, pages 8296--8305, 2020.

\bibitem{alaluf2021hyperstyle}
Yuval Alaluf, Omer Tov, Ron Mokady, Rinon Gal, and Amit~H. Bermano.
\newblock Hyperstyle: Stylegan inversion with hypernetworks for real image
  editing, 2021.

\bibitem{binkowski2018demystifying}
Mikołaj Bińkowski, Dougal~J. Sutherland, Michael Arbel, and Arthur Gretton.
\newblock Demystifying {MMD} {GAN}s.
\newblock In {\em International Conference on Learning Representations}, 2018.

\bibitem{chang2018pairedcyclegan}
Huiwen Chang, Jingwan Lu, Fisher Yu, and Adam Finkelstein.
\newblock Pairedcyclegan: Asymmetric style transfer for applying and removing
  makeup.
\newblock In {\em Proceedings of the IEEE conference on computer vision and
  pattern recognition}, pages 40--48, 2018.

\bibitem{chen2017stylebank}
Dongdong Chen, Lu Yuan, Jing Liao, Nenghai Yu, and Gang Hua.
\newblock Stylebank: An explicit representation for neural image style
  transfer.
\newblock In {\em Proceedings of the IEEE conference on computer vision and
  pattern recognition}, pages 1897--1906, 2017.

\bibitem{chen2018cartoongan}
Yang Chen, Yu-Kun Lai, and Yong-Jin Liu.
\newblock Cartoongan: Generative adversarial networks for photo cartoonization.
\newblock In {\em Proceedings of the IEEE conference on computer vision and
  pattern recognition}, pages 9465--9474, 2018.

\bibitem{choi2018stargan}
Yunjey Choi, Minje Choi, Munyoung Kim, Jung-Woo Ha, Sunghun Kim, and Jaegul
  Choo.
\newblock Stargan: Unified generative adversarial networks for multi-domain
  image-to-image translation.
\newblock In {\em Proceedings of the IEEE conference on computer vision and
  pattern recognition}, pages 8789--8797, 2018.

\bibitem{choi2020stargan}
Yunjey Choi, Youngjung Uh, Jaejun Yoo, and Jung-Woo Ha.
\newblock Stargan v2: Diverse image synthesis for multiple domains.
\newblock In {\em Proceedings of the IEEE/CVF conference on computer vision and
  pattern recognition}, pages 8188--8197, 2020.

\bibitem{creswell2018inverting}
Antonia Creswell and Anil~Anthony Bharath.
\newblock Inverting the generator of a generative adversarial network.
\newblock {\em IEEE transactions on neural networks and learning systems},
  30(7):1967--1974, 2018.

\bibitem{deng2021spatially}
Han Deng, Chu Han, Hongmin Cai, Guoqiang Han, and Shengfeng He.
\newblock Spatially-invariant style-codes controlled makeup transfer.
\newblock In {\em Proceedings of the IEEE/CVF Conference on Computer Vision and
  Pattern Recognition}, pages 6549--6557, 2021.

\bibitem{dinh2021hyperinverter}
Tan~M. Dinh, Anh~Tuan Tran, Rang Nguyen, and Binh-Son Hua.
\newblock Hyperinverter: Improving stylegan inversion via hypernetwork.
\newblock In {\em Proceedings of the IEEE/CVF Conference on Computer Vision and
  Pattern Recognition (CVPR)}, 2022.

\bibitem{gatys2016image}
Leon~A Gatys, Alexander~S Ecker, and Matthias Bethge.
\newblock Image style transfer using convolutional neural networks.
\newblock In {\em Proceedings of the IEEE conference on computer vision and
  pattern recognition}, pages 2414--2423, 2016.

\bibitem{gong2020autotoon}
Julia Gong, Yannick Hold-Geoffroy, and Jingwan Lu.
\newblock Autotoon: Automatic geometric warping for face cartoon generation.
\newblock In {\em Proceedings of the IEEE/CVF Winter Conference on Applications
  of Computer Vision}, pages 360--369, 2020.

\bibitem{goodfellow2014generative}
Ian Goodfellow, Jean Pouget-Abadie, Mehdi Mirza, Bing Xu, David Warde-Farley,
  Sherjil Ozair, Aaron Courville, and Yoshua Bengio.
\newblock Generative adversarial nets.
\newblock {\em Advances in neural information processing systems}, 27, 2014.

\bibitem{gu2019ladn}
Qiao Gu, Guanzhi Wang, Mang~Tik Chiu, Yu-Wing Tai, and Chi-Keung Tang.
\newblock Ladn: Local adversarial disentangling network for facial makeup and
  de-makeup.
\newblock In {\em Proceedings of the IEEE/CVF International Conference on
  Computer Vision}, pages 10481--10490, 2019.

\bibitem{heusel2017gans}
Martin Heusel, Hubert Ramsauer, Thomas Unterthiner, Bernhard Nessler, and Sepp
  Hochreiter.
\newblock Gans trained by a two time-scale update rule converge to a local nash
  equilibrium.
\newblock {\em Advances in neural information processing systems}, 30, 2017.

\bibitem{hu2022style}
Xueqi Hu, Qiusheng Huang, Zhengyi Shi, Siyuan Li, Changxin Gao, Li Sun, and
  Qingli Li.
\newblock Style transformer for image inversion and editing.
\newblock {\em arXiv preprint arXiv:2203.07932}, 2022.

\bibitem{huang2017arbitrary}
Xun Huang and Serge Belongie.
\newblock Arbitrary style transfer in real-time with adaptive instance
  normalization.
\newblock In {\em Proceedings of the IEEE international conference on computer
  vision}, pages 1501--1510, 2017.

\bibitem{huang2018multimodal}
Xun Huang, Ming-Yu Liu, Serge Belongie, and Jan Kautz.
\newblock Multimodal unsupervised image-to-image translation.
\newblock In {\em Proceedings of the European conference on computer vision
  (ECCV)}, pages 172--189, 2018.

\bibitem{harkonen2020ganspace}
Erik Härkönen, Aaron Hertzmann, Jaakko Lehtinen, and Sylvain Paris.
\newblock Ganspace: Discovering interpretable gan controls.
\newblock In {\em Proc. NeurIPS}, 2020.

\bibitem{wikiart}
Mohammed Innat.
\newblock Wiki-art : Visual art encyclopedia.
\newblock \url{www.kaggle.com/ipythonx/wikiart-gangogh-creating-art-gan}, 2020.
\newblock Accessed Jan. 2022.

\bibitem{isola2017image}
Phillip Isola, Jun-Yan Zhu, Tinghui Zhou, and Alexei~A Efros.
\newblock Image-to-image translation with conditional adversarial networks.
\newblock In {\em Proceedings of the IEEE conference on computer vision and
  pattern recognition}, pages 1125--1134, 2017.

\bibitem{jiang2020psgan}
Wentao Jiang, Si Liu, Chen Gao, Jie Cao, Ran He, Jiashi Feng, and Shuicheng
  Yan.
\newblock Psgan: Pose and expression robust spatial-aware gan for customizable
  makeup transfer.
\newblock In {\em Proceedings of the IEEE/CVF Conference on Computer Vision and
  Pattern Recognition}, pages 5194--5202, 2020.

\bibitem{johnson2016perceptual}
Justin Johnson, Alexandre Alahi, and Li Fei-Fei.
\newblock Perceptual losses for real-time style transfer and super-resolution.
\newblock In {\em European conference on computer vision}, pages 694--711.
  Springer, 2016.

\bibitem{Karras2020ada}
Tero Karras, Miika Aittala, Janne Hellsten, Samuli Laine, Jaakko Lehtinen, and
  Timo Aila.
\newblock Training generative adversarial networks with limited data.
\newblock In {\em Proc. NeurIPS}, 2020.

\bibitem{Karras2021}
Tero Karras, Miika Aittala, Samuli Laine, Erik H\"ark\"onen, Janne Hellsten,
  Jaakko Lehtinen, and Timo Aila.
\newblock Alias-free generative adversarial networks.
\newblock In {\em Proc. NeurIPS}, 2021.

\bibitem{karras2019style}
Tero Karras, Samuli Laine, and Timo Aila.
\newblock A style-based generator architecture for generative adversarial
  networks.
\newblock In {\em Proceedings of the IEEE/CVF Conference on Computer Vision and
  Pattern Recognition}, pages 4401--4410, 2019.

\bibitem{karras2020analyzing}
Tero Karras, Samuli Laine, Miika Aittala, Janne Hellsten, Jaakko Lehtinen, and
  Timo Aila.
\newblock Analyzing and improving the image quality of stylegan.
\newblock In {\em Proceedings of the IEEE/CVF Conference on Computer Vision and
  Pattern Recognition}, pages 8110--8119, 2020.

\bibitem{kim2019u}
Junho Kim, Minjae Kim, Hyeonwoo Kang, and Kwanghee Lee.
\newblock U-gat-it: unsupervised generative attentional networks with adaptive
  layer-instance normalization for image-to-image translation.
\newblock {\em arXiv preprint arXiv:1907.10830}, 2019.

\bibitem{kwong2021unsupervised}
Sam Kwong, Jialu Huang, and Jing Liao.
\newblock Unsupervised image-to-image translation via pre-trained stylegan2
  network.
\newblock {\em IEEE Transactions on Multimedia}, 2021.

\bibitem{lee2020freezeg}
Bryan Lee.
\newblock Freeze g.
\newblock \url{http://github.com/bryandlee/FreezeG}, 2020.
\newblock Accessed Jan. 2022.

\bibitem{lee2018diverse}
Hsin-Ying Lee, Hung-Yu Tseng, Jia-Bin Huang, Maneesh Singh, and Ming-Hsuan
  Yang.
\newblock Diverse image-to-image translation via disentangled representations.
\newblock In {\em Proceedings of the European conference on computer vision
  (ECCV)}, pages 35--51, 2018.

\bibitem{li2021anigan}
Bing Li, Yuanlue Zhu, Yitong Wang, Chia-Wen Lin, Bernard Ghanem, and Linlin
  Shen.
\newblock Anigan: Style-guided generative adversarial networks for unsupervised
  anime face generation.
\newblock {\em IEEE Transactions on Multimedia}, 2021.

\bibitem{li2020carigan}
Wenbin Li, Wei Xiong, Haofu Liao, Jing Huo, Yang Gao, and Jiebo Luo.
\newblock Carigan: Caricature generation through weakly paired adversarial
  learning.
\newblock {\em Neural Networks}, 132:66--74, 2020.

\bibitem{liu2021blendgan}
Mingcong Liu, Qiang Li, Zekui Qin, Guoxin Zhang, Pengfei Wan, and Wen Zheng.
\newblock Blendgan: Implicitly gan blending for arbitrary stylized face
  generation.
\newblock {\em Advances in Neural Information Processing Systems}, 34, 2021.

\bibitem{liu2017unsupervised}
Ming-Yu Liu, Thomas Breuel, and Jan Kautz.
\newblock Unsupervised image-to-image translation networks.
\newblock {\em Advances in neural information processing systems}, 30, 2017.

\bibitem{liu2019few}
Ming-Yu Liu, Xun Huang, Arun Mallya, Tero Karras, Timo Aila, Jaakko Lehtinen,
  and Jan Kautz.
\newblock Few-shot unsupervised image-to-image translation.
\newblock In {\em Proceedings of the IEEE/CVF International Conference on
  Computer Vision}, pages 10551--10560, 2019.

\bibitem{Liu_2021_CVPR}
Yahui Liu, Enver Sangineto, Yajing Chen, Linchao Bao, Haoxian Zhang, Nicu Sebe,
  Bruno Lepri, Wei Wang, and Marco De~Nadai.
\newblock Smoothing the disentangled latent style space for unsupervised
  image-to-image translation.
\newblock In {\em Proceedings of the IEEE/CVF Conference on Computer Vision and
  Pattern Recognition (CVPR)}, pages 10785--10794, June 2021.

\bibitem{ma2018invertibility}
Fangchang Ma, Ulas Ayaz, and Sertac Karaman.
\newblock Invertibility of convolutional generative networks from partial
  measurements.
\newblock {\em Advances in Neural Information Processing Systems}, 31, 2018.

\bibitem{nizan2020breaking}
Ori Nizan and Ayellet Tal.
\newblock Breaking the cycle-colleagues are all you need.
\newblock In {\em Proceedings of the IEEE/CVF Conference on Computer Vision and
  Pattern Recognition}, pages 7860--7869, 2020.

\bibitem{park2020cut}
Taesung Park, Alexei~A. Efros, Richard Zhang, and Jun-Yan Zhu.
\newblock Contrastive learning for unpaired image-to-image translation.
\newblock In {\em European Conference on Computer Vision}, 2020.

\bibitem{paszke2019pytorch}
Adam Paszke, Sam Gross, Francisco Massa, Adam Lerer, James Bradbury, Gregory
  Chanan, Trevor Killeen, Zeming Lin, Natalia Gimelshein, Luca Antiga, et~al.
\newblock Pytorch: An imperative style, high-performance deep learning library.
\newblock {\em Advances in neural information processing systems}, 32, 2019.

\bibitem{pinkney2020resolution}
Justin~NM Pinkney and Doron Adler.
\newblock Resolution dependent gan interpolation for controllable image
  synthesis between domains.
\newblock {\em arXiv preprint arXiv:2010.05334}, 2020.

\bibitem{richardson2021encoding}
Elad Richardson, Yuval Alaluf, Or Patashnik, Yotam Nitzan, Yaniv Azar, Stav
  Shapiro, and Daniel Cohen-Or.
\newblock Encoding in style: a stylegan encoder for image-to-image translation.
\newblock In {\em Proceedings of the IEEE/CVF Conference on Computer Vision and
  Pattern Recognition}, pages 2287--2296, 2021.

\bibitem{sanakoyeu2018style}
Artsiom Sanakoyeu, Dmytro Kotovenko, Sabine Lang, and Bjorn Ommer.
\newblock A style-aware content loss for real-time hd style transfer.
\newblock In {\em proceedings of the European conference on computer vision
  (ECCV)}, pages 698--714, 2018.

\bibitem{shen2018neural}
Falong Shen, Shuicheng Yan, and Gang Zeng.
\newblock Neural style transfer via meta networks.
\newblock In {\em Proceedings of the IEEE Conference on Computer Vision and
  Pattern Recognition}, pages 8061--8069, 2018.

\bibitem{shen2021closed}
Yujun Shen and Bolei Zhou.
\newblock Closed-form factorization of latent semantics in gans.
\newblock In {\em Proceedings of the IEEE/CVF Conference on Computer Vision and
  Pattern Recognition}, pages 1532--1540, 2021.

\bibitem{Sheng_2018_CVPR}
Lu Sheng, Ziyi Lin, Jing Shao, and Xiaogang Wang.
\newblock Avatar-net: Multi-scale zero-shot style transfer by feature
  decoration.
\newblock In {\em Proceedings of the IEEE Conference on Computer Vision and
  Pattern Recognition (CVPR)}, June 2018.

\bibitem{shi2019warpgan}
Yichun Shi, Debayan Deb, and Anil~K Jain.
\newblock Warpgan: Automatic caricature generation.
\newblock In {\em Proceedings of the IEEE/CVF Conference on Computer Vision and
  Pattern Recognition}, pages 10762--10771, 2019.

\bibitem{song2021agilegan}
Guoxian Song, Linjie Luo, Jing Liu, Wan-Chun Ma, Chunpong Lai, Chuanxia Zheng,
  and Tat-Jen Cham.
\newblock Agilegan: stylizing portraits by inversion-consistent transfer
  learning.
\newblock {\em ACM Transactions on Graphics (TOG)}, 40(4):1--13, 2021.

\bibitem{su2021mangagan}
Hao Su, Jianwei Niu, Xuefeng Liu, Qingfeng Li, Jiahe Cui, and Ji Wan.
\newblock Mangagan: Unpaired photo-to-manga translation based on the
  methodology of manga drawing.
\newblock In {\em Proceedings of the AAAI Conference on Artificial
  Intelligence}, volume~35, pages 2611--2619, 2021.

\bibitem{10.1145/3450626.3459838}
Omer Tov, Yuval Alaluf, Yotam Nitzan, Or Patashnik, and Daniel Cohen-Or.
\newblock Designing an encoder for stylegan image manipulation.
\newblock {\em ACM Trans. Graph.}, 40(4), jul 2021.

\bibitem{wang2021HFGI}
Tengfei Wang, Yong Zhang, Yanbo Fan, Jue Wang, and Qifeng Chen.
\newblock High-fidelity gan inversion for image attribute editing.
\newblock In {\em Proceedings of the IEEE/CVF Conference on Computer Vision and
  Pattern Recognition (CVPR)}, 2022.

\bibitem{wang2018high}
Ting-Chun Wang, Ming-Yu Liu, Jun-Yan Zhu, Andrew Tao, Jan Kautz, and Bryan
  Catanzaro.
\newblock High-resolution image synthesis and semantic manipulation with
  conditional gans.
\newblock In {\em Proceedings of the IEEE conference on computer vision and
  pattern recognition}, pages 8798--8807, 2018.

\bibitem{wang2020learning}
Xinrui Wang and Jinze Yu.
\newblock Learning to cartoonize using white-box cartoon representations.
\newblock In {\em Proceedings of the IEEE/CVF Conference on Computer Vision and
  Pattern Recognition}, pages 8090--8099, 2020.

\bibitem{wu2021stylespace}
Zongze Wu, Dani Lischinski, and Eli Shechtman.
\newblock Stylespace analysis: Disentangled controls for stylegan image
  generation.
\newblock In {\em Proceedings of the IEEE/CVF Conference on Computer Vision and
  Pattern Recognition}, pages 12863--12872, 2021.

\bibitem{wu2022stylealign}
Zongze Wu, Yotam Nitzan, Eli Shechtman, and Dani Lischinski.
\newblock Stylealign: Analysis and applications of aligned style{GAN} models.
\newblock In {\em International Conference on Learning Representations}, 2022.

\bibitem{xu2021generative}
Yinghao Xu, Yujun Shen, Jiapeng Zhu, Ceyuan Yang, and Bolei Zhou.
\newblock Generative hierarchical features from synthesizing images.
\newblock In {\em CVPR}, 2021.

\bibitem{yang2022pastiche}
Shuai Yang, Liming Jiang, Ziwei Liu, and Chen~Change Loy.
\newblock Pastiche master: Exemplar-based high-resolution portrait style
  transfer.
\newblock In {\em Proceedings of the IEEE/CVF Conference on Computer Vision and
  Pattern Recognition}, pages 7693--7702, 2022.

\bibitem{yang2022Vtoonify}
Shuai Yang, Liming Jiang, Ziwei Liu, and Chen~Change Loy.
\newblock Vtoonify: Controllable high-resolution portrait video style transfer.
\newblock {\em ACM Transactions on Graphics (TOG)}, 41(6):1--15, 2022.

\bibitem{yu2015lsun}
Fisher Yu, Ari Seff, Yinda Zhang, Shuran Song, Thomas Funkhouser, and Jianxiong
  Xiao.
\newblock Lsun: Construction of a large-scale image dataset using deep learning
  with humans in the loop.
\newblock {\em arXiv preprint arXiv:1506.03365}, 2015.

\bibitem{zhang2018perceptual}
Richard Zhang, Phillip Isola, Alexei~A Efros, Eli Shechtman, and Oliver Wang.
\newblock The unreasonable effectiveness of deep features as a perceptual
  metric.
\newblock In {\em CVPR}, 2018.

\bibitem{zheng2021spatiallycorrelative}
Chuanxia Zheng, Tat-Jen Cham, and Jianfei Cai.
\newblock The spatially-correlative loss for various image translation tasks.
\newblock In {\em Proceedings of the IEEE Conference on Computer Vision and
  Pattern Recognition}, 2021.

\bibitem{zhu2020indomain}
Jiapeng Zhu, Yujun Shen, Deli Zhao, and Bolei Zhou.
\newblock In-domain gan inversion for real image editing.
\newblock In {\em Proceedings of European Conference on Computer Vision
  (ECCV)}, 2020.

\bibitem{zhu2017unpaired}
Jun-Yan Zhu, Taesung Park, Phillip Isola, and Alexei~A Efros.
\newblock Unpaired image-to-image translation using cycle-consistent
  adversarial networks.
\newblock In {\em Proceedings of the IEEE international conference on computer
  vision}, pages 2223--2232, 2017.

\bibitem{zhu2017toward}
Jun-Yan Zhu, Richard Zhang, Deepak Pathak, Trevor Darrell, Alexei~A Efros,
  Oliver Wang, and Eli Shechtman.
\newblock Toward multimodal image-to-image translation.
\newblock {\em Advances in neural information processing systems}, 30, 2017.

\end{thebibliography}
}

\ifarxiv
\clearpage
\appendix
\section{Evaluation protocol}
\noindent\textbf{Comparison with unconditional GANs based method.}
We adopt Fr\'echet Inception Distance (FID)\cite{heusel2017gans} and Kernel Inception Distance (KID) \cite{binkowski2018demystifying} to measure the generation quality and diversity of generated images.
FID and KID are computed between 50K generated images and the entire training samples.
We use a modified version of  Perceptual Smoothness (PS) \cite{Liu_2021_CVPR} to measure the smoothness of interpolation between different domain features.
Instead of the style code which is used in the original paper \cite{Liu_2021_CVPR}, we use $\mathbf z \in \mathcal Z$ for target interpolation latent.
Note that this modification is due to the architectural difference.
Same as FID and KID, 50K samples are used to compute PS.
In order to ensure that the implementation difference does not affect performance, we compare all methods above the official Pytorch \cite{paszke2019pytorch} implementation of StyleGAN2-ADA\footnote{\url{https://github.com/NVlabs/stylegan2-ada-pytorch}} \cite{Karras2020ada}.

\noindent\textbf{Comparison on domain translation method.}
We use FID and KID to evaluate generated images.
20K images are randomly sampled from the source domain.
For our approach, we project source domain images to $\mathbf z \in \mathcal Z$ and provide them to the target model.
For the other domain translation methods, source domain images and corresponding randomly sampled style latent codes are used to generate images.
Note that 20K generated images and the entire target domain images are used for evaluation.

\section{Additional results}
\label{sec:results}
\noindent\textbf{Evaluation on anchor point $n_{anch}$.}
We evaluate the proposed method using different anchor points in FFHQ $\rightarrow$ Metfaces setting.
We train our model from scratch 10 times and report Perceptual Smoothness (PS) \cite{Liu_2021_CVPR}, FID \cite{heusel2017gans}, and KID \cite{binkowski2018demystifying} in Table~\ref{table:noise_exp}.
The anchor point $n_{anch}$ is randomly sampled from the Gaussian distribution for each experiment.

\noindent\textbf{Noise interplation.} 
We provide additional noise interpolation results of the proposed method on FFHQ $\rightarrow$ MetFaces (Figure~\ref{fig:ours_metface}), FFHQ $\rightarrow$ AAHQ (Figure~\ref{fig:ours_aahq}) and LSUN Church $\rightarrow$ WikiArt Cityscape (Figure~\ref{fig:ours_wikiart}).

\setlength{\tabcolsep}{8pt}
\begin{table}[ht]
\begin{center}
\small
\begin{tabular}{c|ccc}
\toprule
Setting        & \multicolumn{3}{c}{FFHQ $\rightarrow$ MetFaces} \\ \midrule
$\alpha$  &  PS& FID & \vtop{\hbox{KID} \hbox{\tiny{($\times 10^3)$}}}  \\
\midrule

1     & \multirow{2}{*}{0.884 $\pm$ 0.04} & 38.69 $\pm$ 3.23 & 14.36 $\pm$ 1.93    \\
0     &                                   & 20.08 $\pm$ 0.30 & \,\;3.54 $\pm$ 0.37     \\

\bottomrule
\end{tabular}
\caption{Experiment on different anchor point $n_{anch}$.
We report the mean and standard deviation of metrics over 10 runs.}
\label{table:noise_exp}
\end{center}
\end{table}
\setlength{\tabcolsep}{0.5pt}

\noindent\textbf{Comparison with unconditional GANs based method.}
An additional qualitative comparison of controlling preserved source features is shown in Figure ~\ref{fig:compare_metface},~\ref{fig:compare_aahq}, and ~\ref{fig:compare_wikiart}.
Freeze G \cite{lee2020freezeg} that requires new training for each source degree shows an inconsistent transition of the preserved source features.
Layer-swap \cite{pinkney2020resolution} and UI2I StyleGAN2 \cite{kwong2021unsupervised} that convert weights of a source model also show the inconsistent transition.
Specifically, unnatural color transitions from the source domain are observed in Figure ~\ref{fig:compare_metface}. Additionally, several artifacts and changes in the human identity are observed in Figure ~\ref{fig:compare_aahq}.
We believe that this phenomenon occurs due to the long training time of the target model (\textit{e.g.} FFHQ $\rightarrow$ AAHQ are trained for 12000K images).
The long training time causes more changes in the target model weights, and this may disturb the combined models to generate realistic images.
For example, the identity changes seen in the result of layer swap (Figure~\ref{fig:compare_aahq}) seem to be caused by a large change in the mapping function that transforms $\mathbf z \in \mathcal Z$ to $\mathbf w \in \mathcal W$.
The color transition problems and inconsistent transition are less observable in LSUN Church $\rightarrow$ WikiArt Cityscape, due to the artistic target dataset and the spatial difference between the source and target domain, respectively.
Nevertheless, these methods require models for each degree of preserved source features, while  the proposed method can control in a single model.

\begin{figure}[t!]
\centering
\includegraphics[width=1\linewidth]{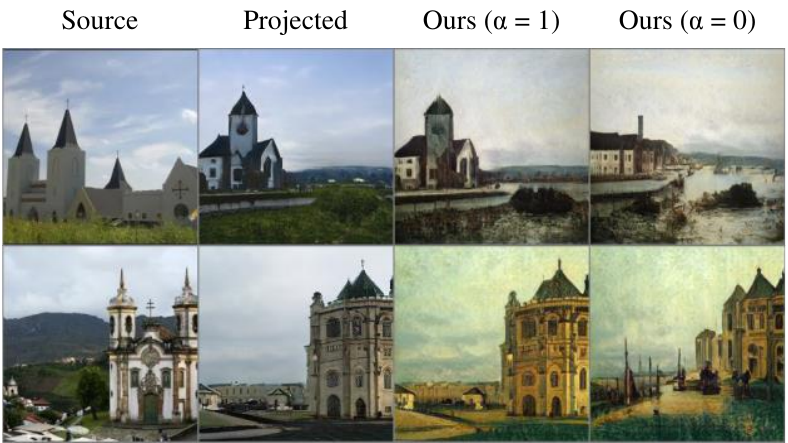}
   \caption{Domain translation results for Church $\rightarrow$ Cityscape. Incorrectly projected images cause our method to generate uncorrelated target images with source.}
\label{fig:project_fail}
\end{figure}

\noindent\textbf{Comparison on domain translation method.}
An additional qualitative comparison of controlling preserved source features are shown in Figure ~\ref{fig:compare_DT_Metfaces},~\ref{fig:compare_DT_AAHQ}.
The latent inversion method is only used for our approach.
Modified version of the inversion method in StyleGAN2 \cite{karras2019style} is used.
We embed real images into the $\mathcal Z$ space of the source model with truncation psi of 0.7 following StyleAlign \cite{wu2022stylealign}.
In the comparison, we use the exact same latent code obtained by the inversion method for the target model.
However, please note that our method can also be multimodal like MUNIT \cite{huang2018multimodal} and StarGAN-v2 \cite{choi2020stargan} by combining early latent code from the projected latent code with the late latent code from the others \cite{wu2022stylealign}.

\noindent\textbf{Latent inversion failure cases.}
Projecting images into the $\mathcal Z$ space of the StyleGAN often fails to accurately reconstruct original images when the dataset becomes larger and more diverse.
The inversion and translated results on Church $\rightarrow$ Cityscape are shown in Figure~\ref{fig:DT_church_fail}.
The result shows a strong correlation between projected and translated images.
However, the incorrectly acquired latent codes lead to uncorrelated target domain images.
It would be interesting to integrate our method well with the inversion method for other spaces (\textit{e.g.} $\mathcal Z+, \mathcal W$, and $\mathcal W+$), or to improve the performance of the inversion method for $\mathcal Z$ space.

\section{Latent modulation}
Recently, several works \cite{harkonen2020ganspace,shen2021closed,wu2021stylespace} observe that StyleGAN can effectively adjust semantic attributes of images by modulating latent codes in interpretable directions.
Additionally, StyleSpace \cite{wu2021stylespace} revealed that the $\mathcal{S}$ space is the most disentangled among the three latent spaces $\mathcal{Z}$, $\mathcal{W}$, and $\mathcal{S}$ of StyleGAN \cite{karras2019style,karras2020analyzing}, and it is possible to change various semantic attributes of generated images just by adjusting a value of the single dimension of $\mathcal S$.
Based on this observation, we examine latent modulation effects on our proposed method. The latent modulation effects on different interpolation weights are shown in Figure~\ref{fig:latent_metface}, \ref{fig:latent_aahq}, and \ref{fig:latent_wikiart}.
The latent modulation effects of the source model are highly aligned in anchored subspace ($\alpha = 1$).
As $\alpha$ decreases, some latent modulation effects remain, while the rest gradually weakens or disappears.
This phenomenon may occur as the preserved source features gradually vanish.

\section{Comparison with SmoothingLatentSpace}
Our approach allows smooth interpolation between the source and target features in the transfer-learned model. 
We additionally compare our approach with SmoothingLatentSpace \cite{Liu_2021_CVPR} which tries to smooth the interpolation between the source and target domain.
For SmoothingLatentSpace, We interpolate latent codes from source images $\mathbf s_{s}$ and randomly sampled noise $\mathbf s_{rand}$, $\alpha \cdot \mathbf s_{s} + (1-\alpha) \cdot \mathbf s_{rand}$,  and generate target images with content from source images and interpolated latent codes.
Figure \ref{fig:compare_DT_smoothing_metfaces} and \ref{fig:compare_DT_smoothing_AAHQ} show interpolation results between the source and target features.
The results show that SmoothingLatentSpace frequently generates severe artifacts during the interpolation between latent codes from source images and randomly sampled noise.
In addition, compared to our method, SmoothingLatentSpace generates less smooth interpolation results.

\section{Limitations}
Despite our method achieved compelling results, it is not without limitations.
Although our method is easily applicable to StyleGAN 1 \cite{karras2019style} and 2 \cite{karras2020analyzing}, it is hard to directly incorporate our method into architectures that do not contain the noise input such as StyleGAN3 \cite{Karras2021} which removed the noise input to achieve equivariances.
Second, inversion methods for $\mathcal Z$ space cannot accurately reconstruct finer details of real images, which interferes with the consistency between the source and target images in domain translation.
For example, the results in Figure 6 in the main paper
show slight changes in the face identity due to inaccurately obtained latent codes.
Additionally, this phenomenon is exacerbated when the dataset becomes larger and more diverse.
As shown in Figure~\ref{fig:project_fail}, the latent inversion method causes significant changes in overall reconstructed images in LSUN church, which leads our method to generate target images uncorrelated with source images.
In the future, it might be interesting to design inversion methods that overcome the above issues.

\section{Broader impact}
Translating one image to other domains has received tremendous attention from the community and has been used in a variety of applications.
In addition to generating various images from one image (multimodal), it is also very important to determine how much of the source features are preserved.
For example, users may obtain results in which the desired degree of characteristics is preserved in the applications.
As such, we see great potential for our technology to be utilized in various applications.

However, since our method is based on data-driven generative modeling, it faces various ethical issues arising from bias in the training data.
For example, a target model fine-tuned from a source model pre-trained on FFHQ tends to generate more light-skinned images than dark-skinned ones.
In addition, the phenomenon of changing dark-skinned images to light-colored skin was also observed.
At a time when data-driven modeling is getting a lot of attention, the community needs a lot of effort and discussion about data bias.



\begin{figure*}[t!]
\centering
\includegraphics[width=0.9\linewidth]{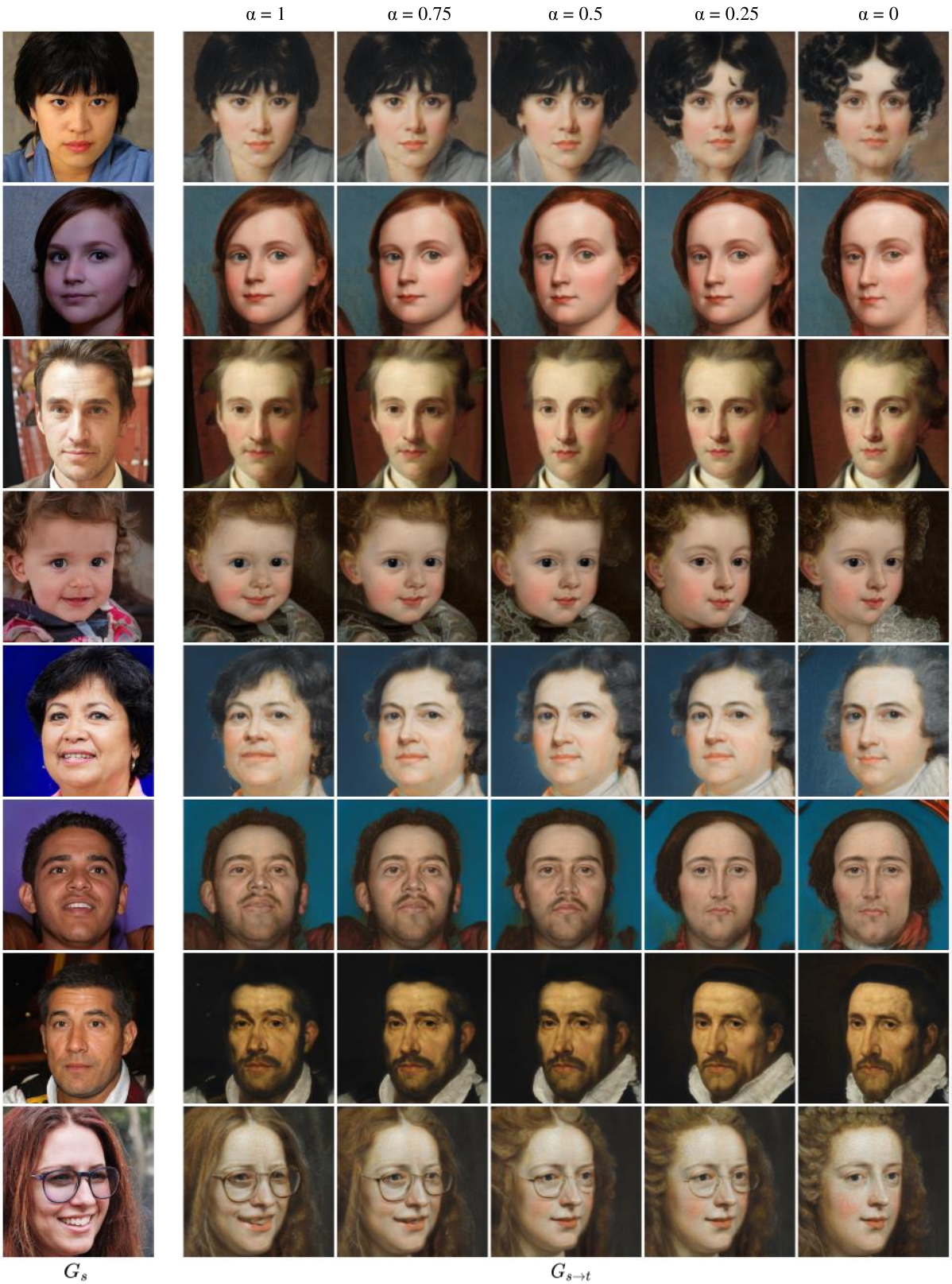}
   \caption{\textbf{[FFHQ $\rightarrow$ MetFaces]}
   Visualizing the effects of the noise interpolation.
   The interpolation weight $\alpha$ is presented above each column.}
\label{fig:ours_metface}
\end{figure*}

\begin{figure*}[t!]
\centering
\includegraphics[width=0.9\linewidth]{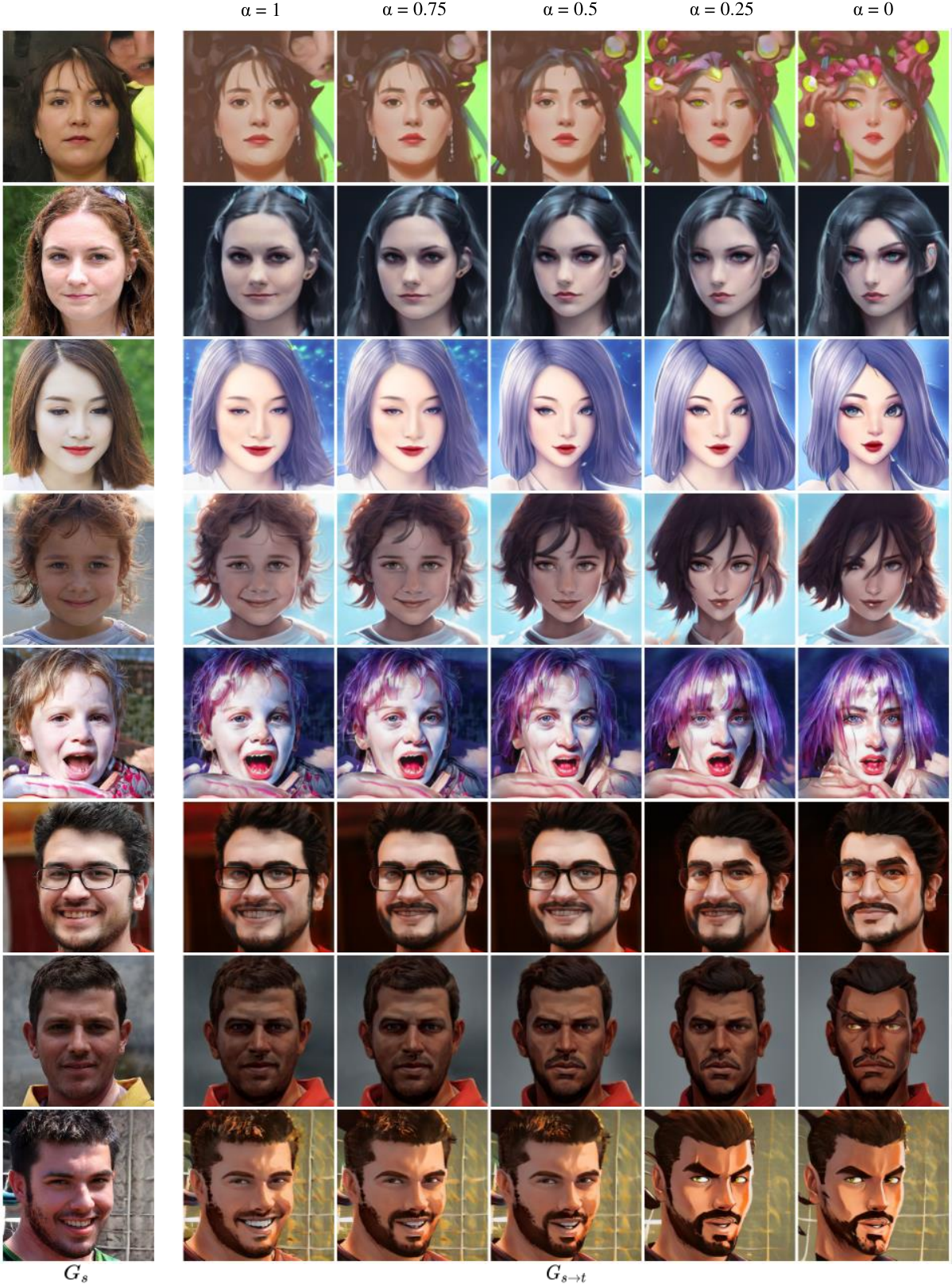}
   \caption{\textbf{[FFHQ $\rightarrow$ AAHQ]}
   Visualizing the effects of the noise interpolation.
   The interpolation weight $\alpha$ is presented above each column.}
\label{fig:ours_aahq}
\end{figure*}

\begin{figure*}[t!]
\centering
\includegraphics[width=0.9\linewidth]{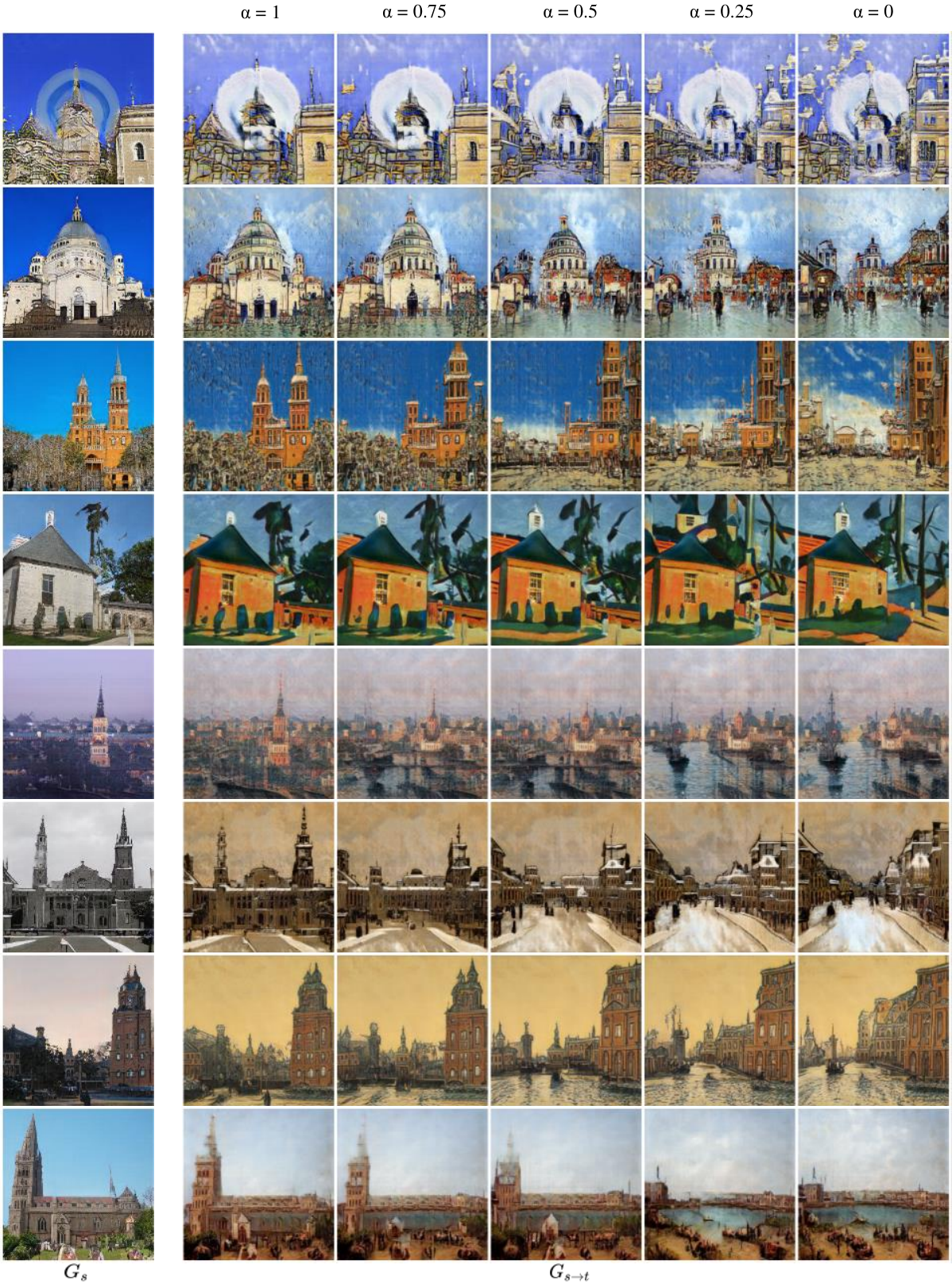}
   \caption{\textbf{[Church $\rightarrow$ Cityscape]}
   Visualizing the effects of the noise interpolation.
   The interpolation weight $\alpha$ is presented above each column.}
\label{fig:ours_wikiart}
\end{figure*}

\begin{figure*}[t!]
\centering
\includegraphics[width=0.9\linewidth]{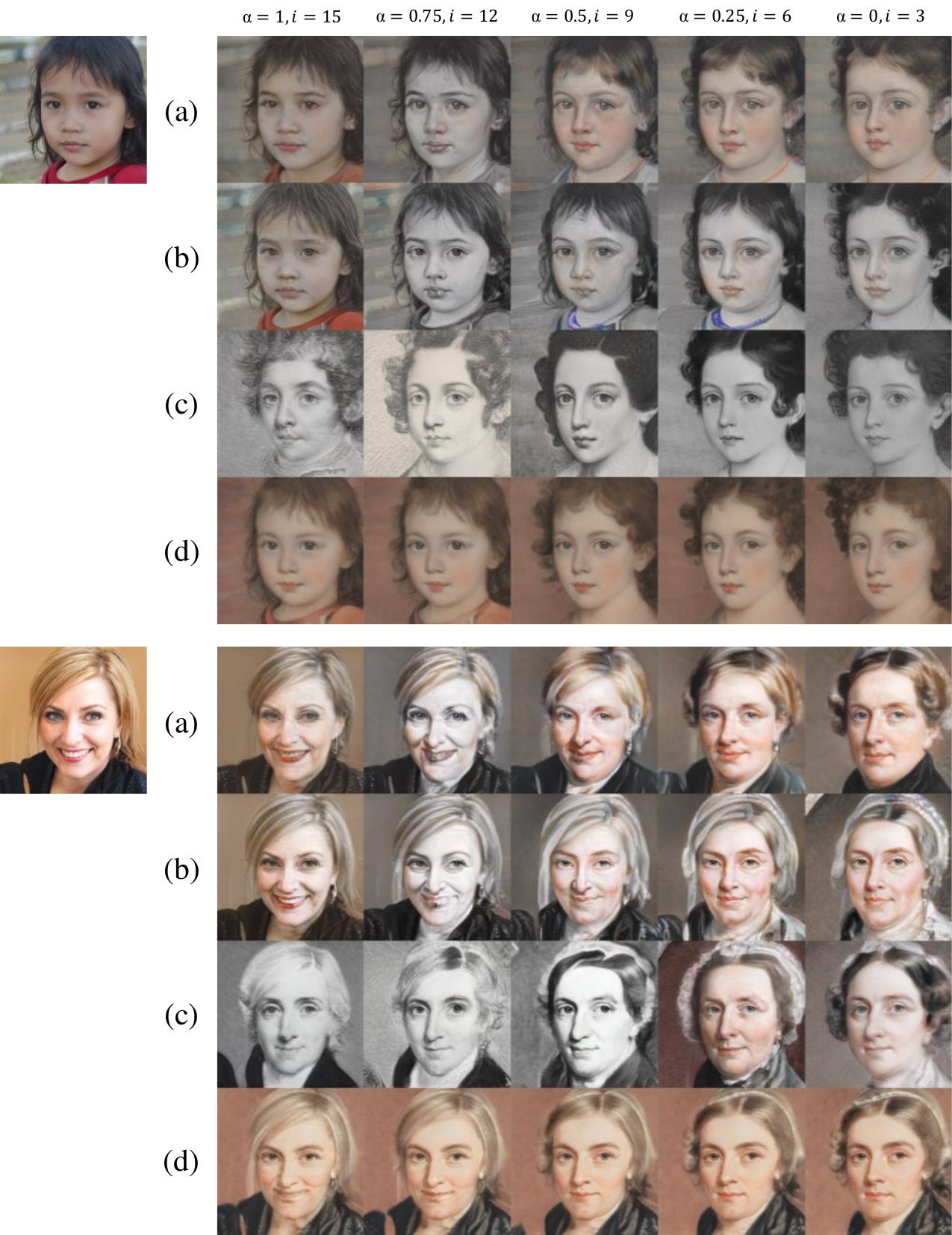}
   \caption{\textbf{[FFHQ $\rightarrow$ MetFaces]} Qualitative comparison on controlling  preserved source features: (a) Layer-swap \cite{pinkney2020resolution}, (b) UI2I StyleGAN2 \cite{kwong2021unsupervised}, (c) Freeze G \cite{lee2020freezeg}, (d) ours.
   The interpolation weight $\alpha$ and swap / freeze layer $i$ are presented above each column.}
\label{fig:compare_metface}
\end{figure*}

\begin{figure*}[t!]
\centering
\includegraphics[width=0.9\linewidth]{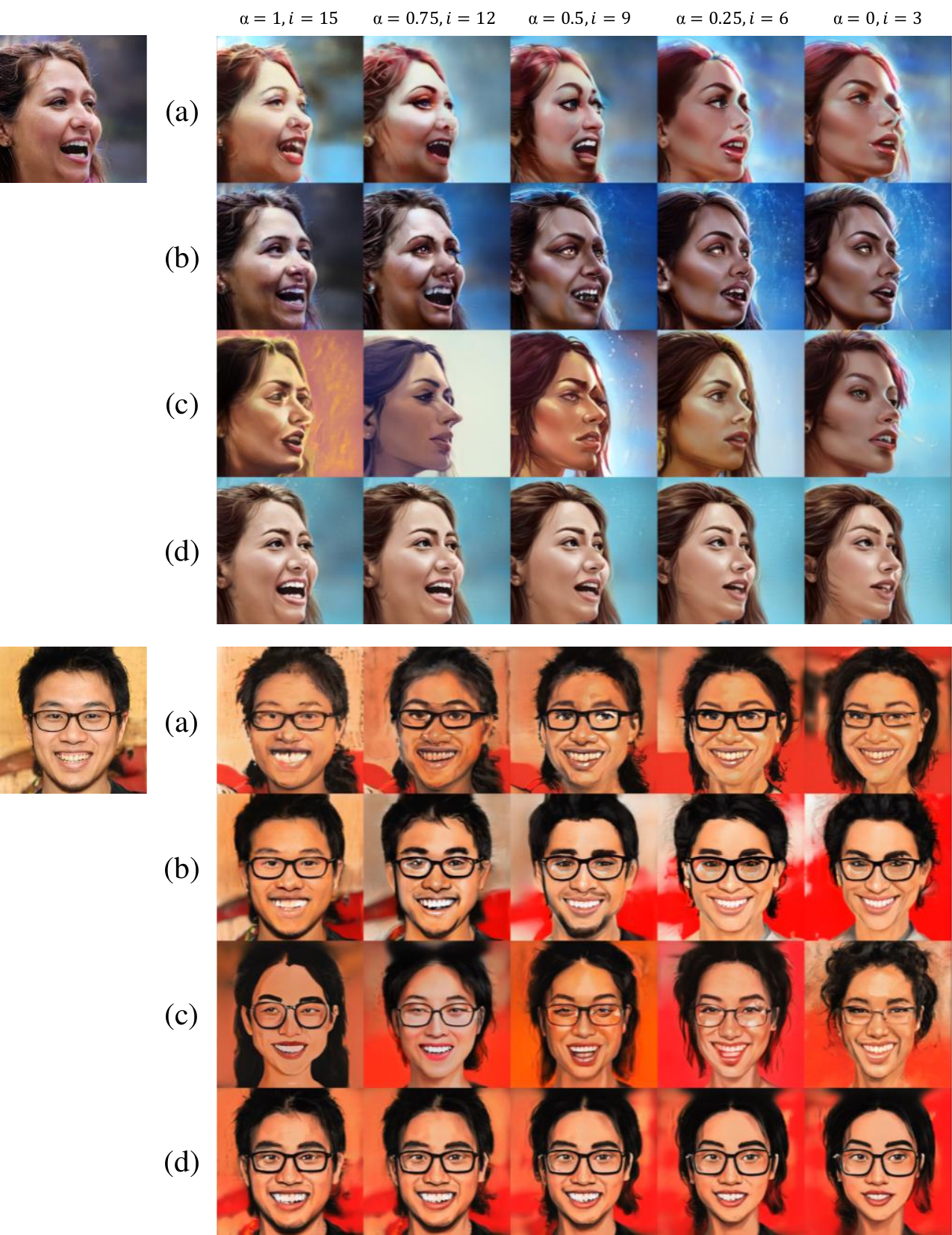}
   \caption{\textbf{[FFHQ $\rightarrow$ AAHQ]} Qualitative comparison on controlling  preserved source features: (a) Layer-swap \cite{pinkney2020resolution}, (b) UI2I StyleGAN2 \cite{kwong2021unsupervised}, (c) Freeze G \cite{lee2020freezeg}, (d) ours.
   The interpolation weight $\alpha$ and swap / freeze layer $i$ are presented above each column.}
\label{fig:compare_aahq}
\end{figure*}

\begin{figure*}[t!]
\centering
\includegraphics[width=0.9\linewidth]{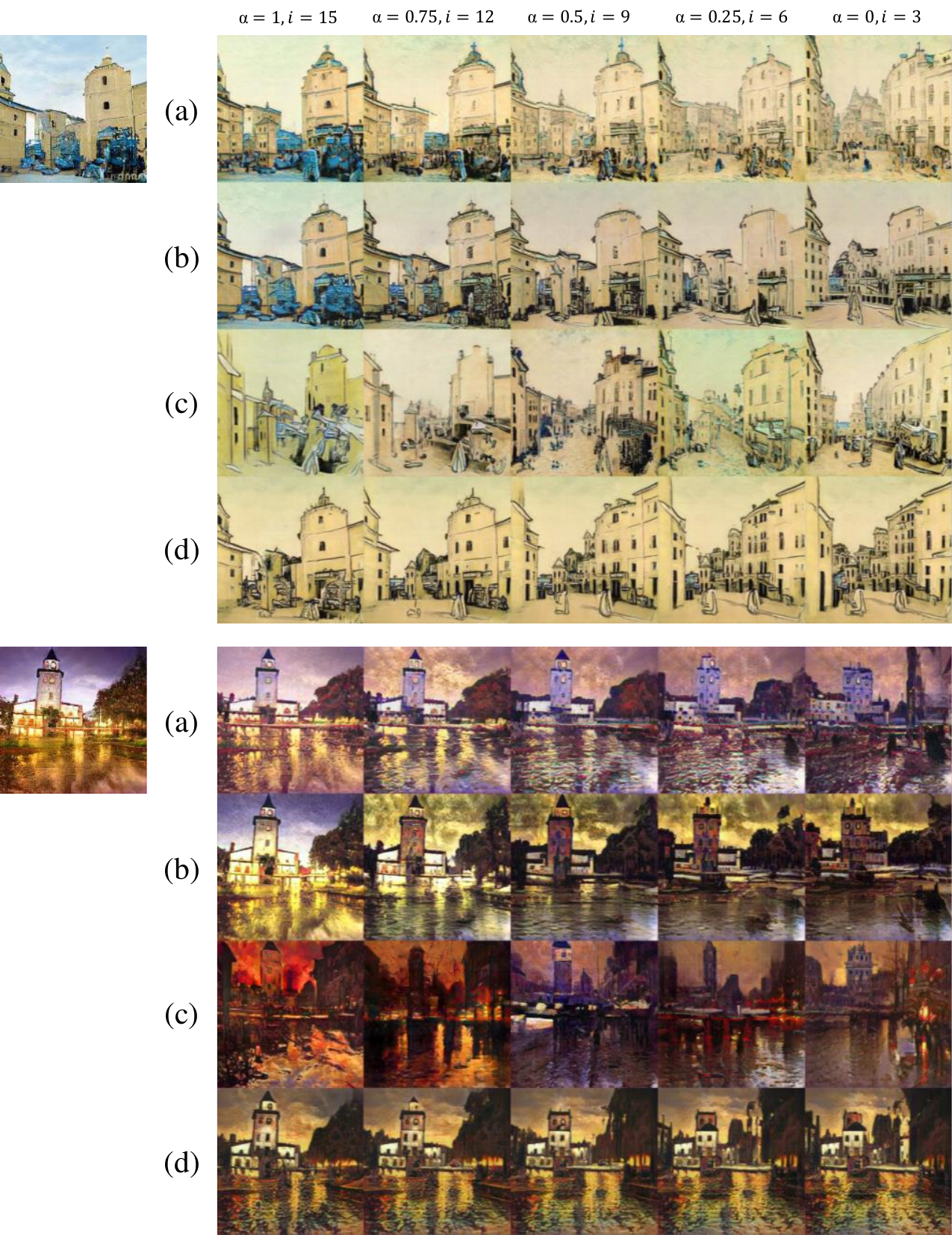}
   \caption{\textbf{[Church $\rightarrow$ Cityscape]} Qualitative comparison on controlling  preserved source features: (a) Layer-swap \cite{pinkney2020resolution}, (b) UI2I StyleGAN2 \cite{kwong2021unsupervised}, (c) Freeze G \cite{lee2020freezeg}, (d) ours.
   The interpolation weight $\alpha$ and swap / freeze layer $i$ are presented above each column.}
\label{fig:compare_wikiart}
\end{figure*}

\begin{figure*}[t!]
\centering
\includegraphics[width=0.9\linewidth]{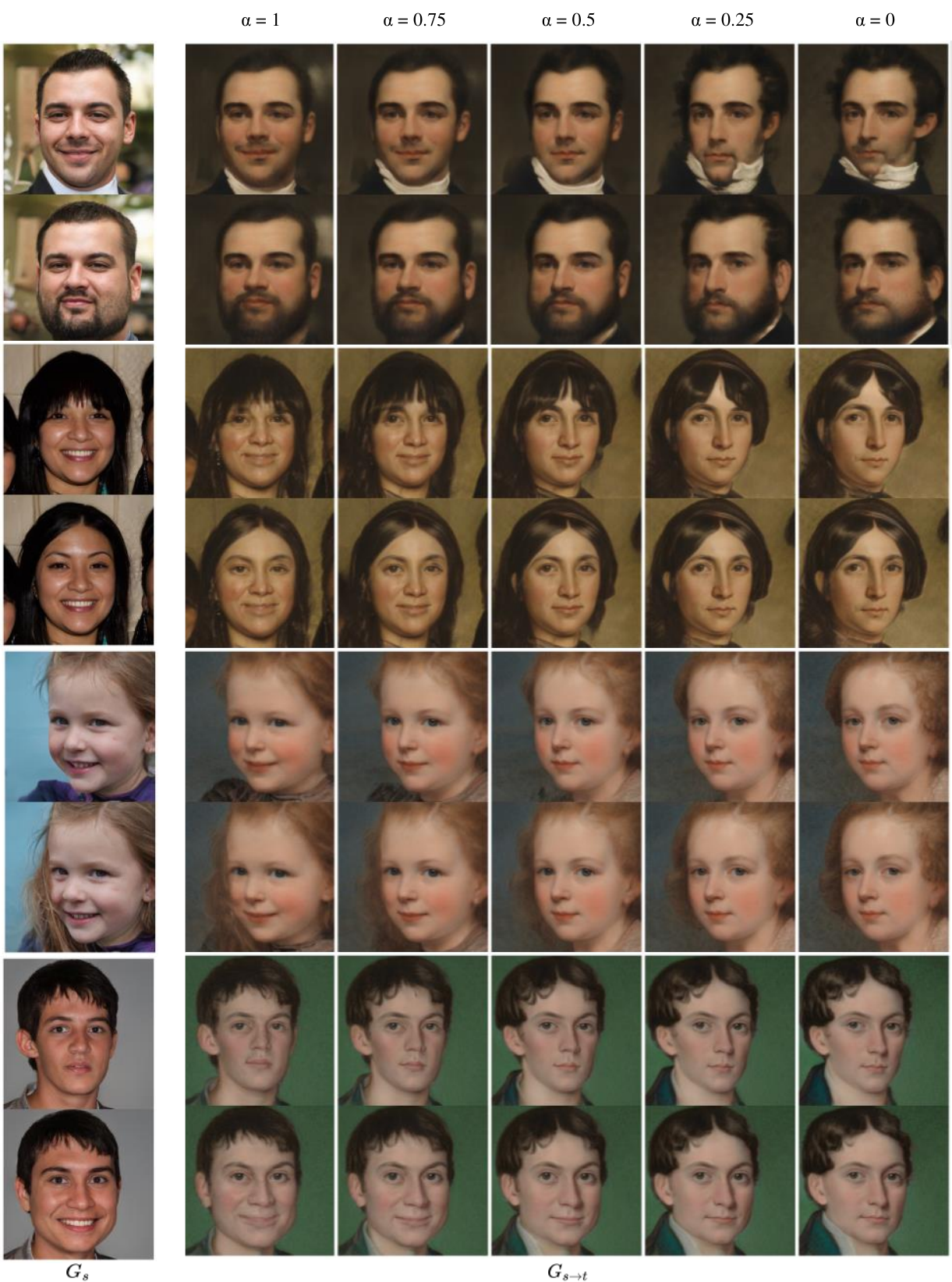}
   \caption{\textbf{[FFHQ $\rightarrow$ MetFaces]}
   Visualizing the effects of the latent modulation on different interpolation weight.
   Each of the two adjacent columns is the result of modulating the latent in a different direction ($+/-$).
   The interpolation weight $\alpha$ is presented above each column.}
\label{fig:latent_metface}
\end{figure*}

\begin{figure*}[t!]
\centering
\includegraphics[width=0.9 \linewidth]{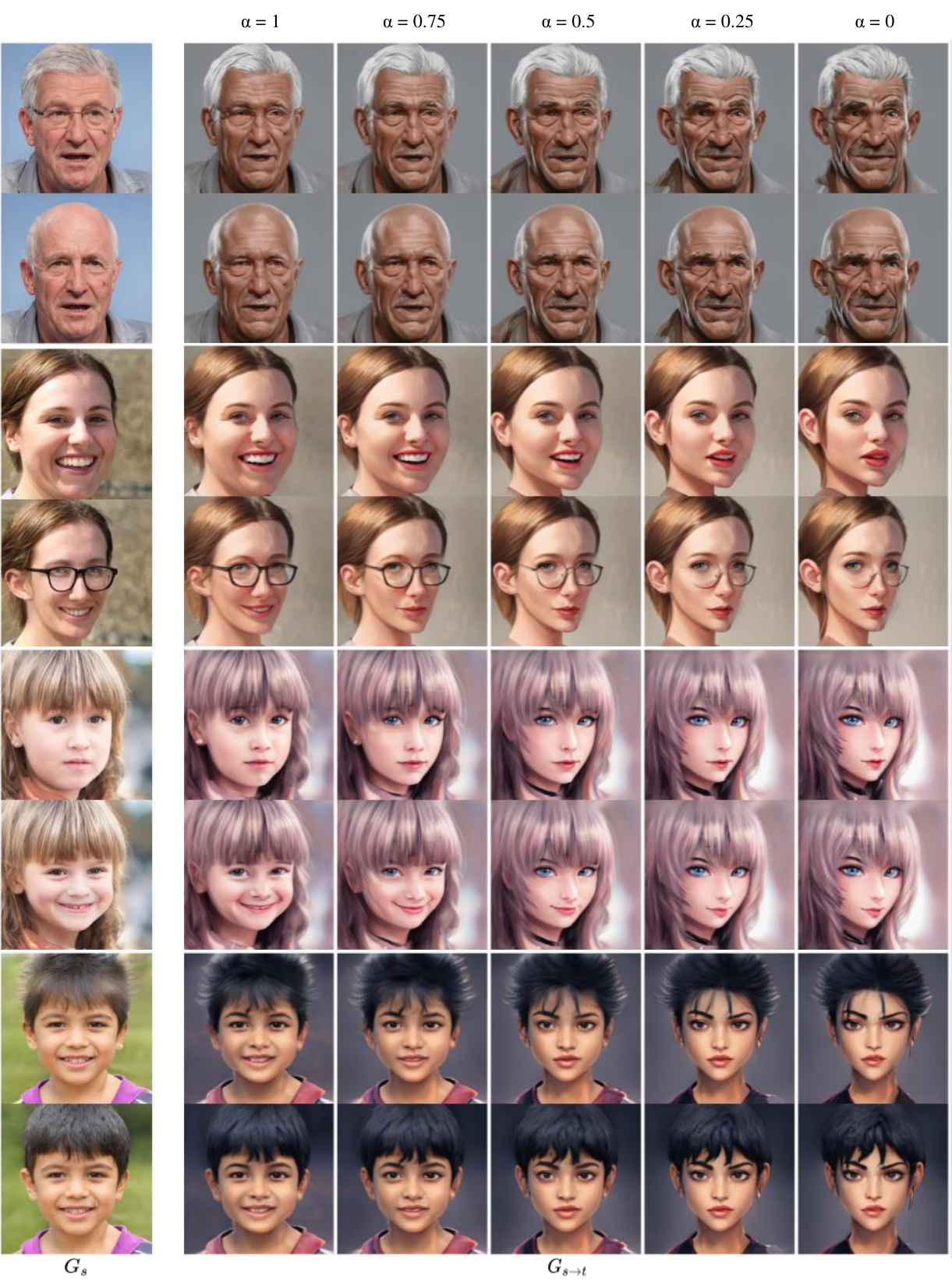}
   \caption{\textbf{[FFHQ $\rightarrow$ AAHQ]}
   Visualizing the effects of the latent modulation on different interpolation weight.
   Each of the two adjacent columns is the result of modulating the latent in a different direction ($+/-$).
   The interpolation weight $\alpha$ is presented above each column.}
\label{fig:latent_aahq}
\end{figure*}

\begin{figure*}[t!]
\centering
\includegraphics[width=0.9 \linewidth]{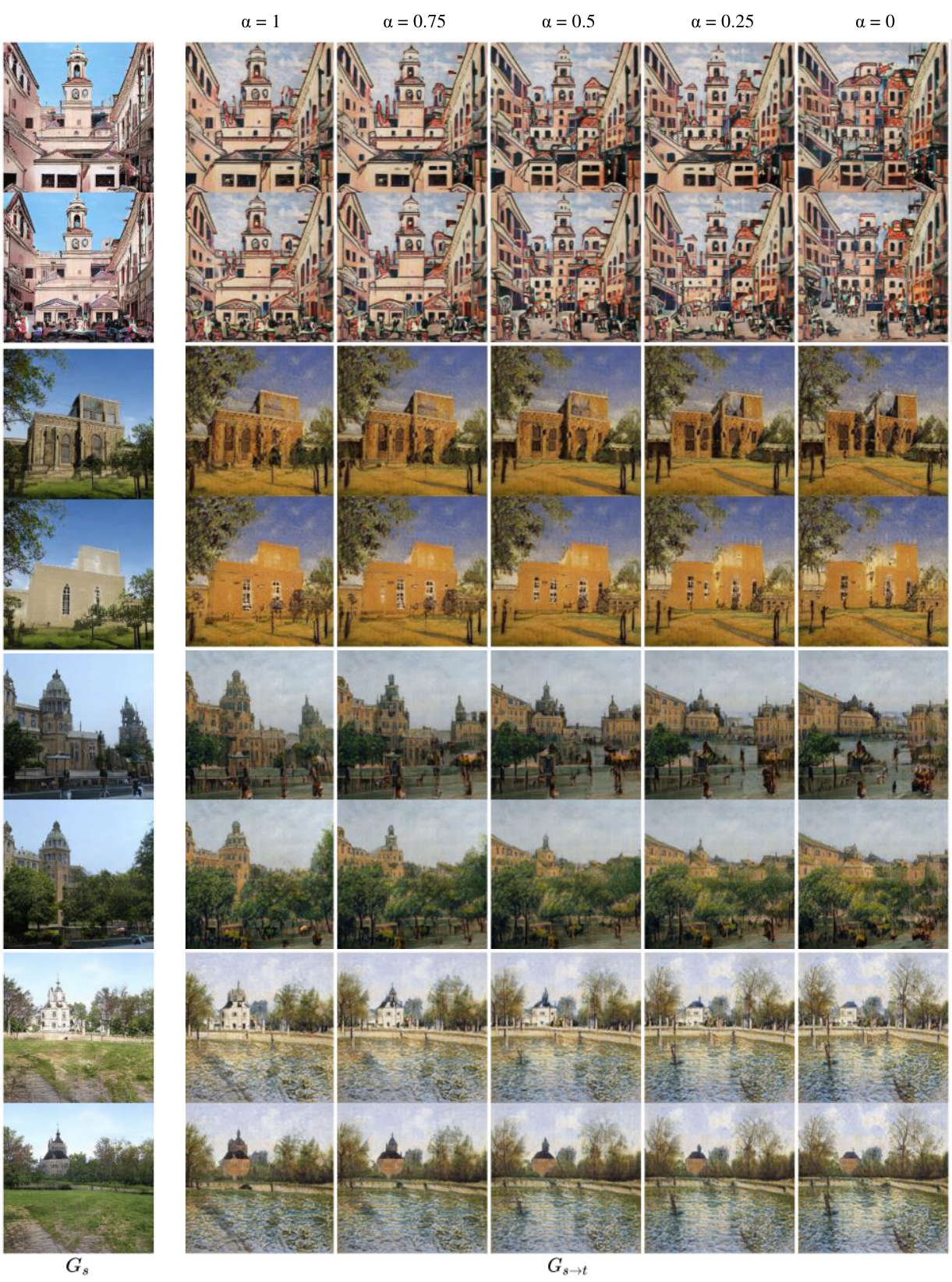}
   \caption{\textbf{[Church $\rightarrow$ Cityscape]}
   Visualizing the effects of the latent modulation on different interpolation weight.
   Each of the two adjacent columns is the result of modulating the latent in a different direction ($+/-$).
   The interpolation weight $\alpha$ is presented above each column.}
\label{fig:latent_wikiart}
\end{figure*}

\begin{figure*}[t!]
\centering
\includegraphics[width=0.82\linewidth]{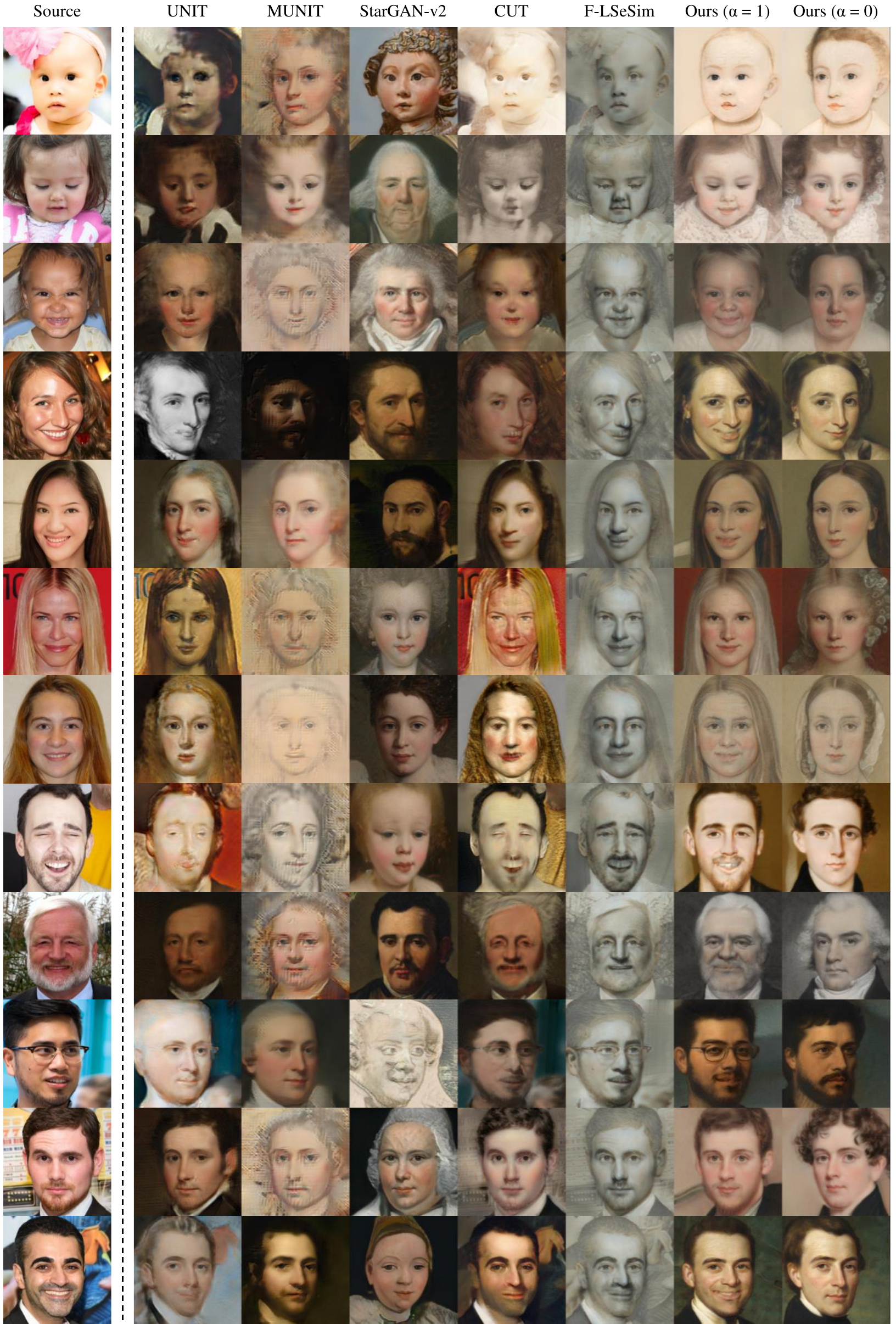}
   \caption{\textbf{[FFHQ $\rightarrow$ MetFaces]}
   Qualitative comparison on domain translation.
   Our method is not only qualitatively best, but also can control source features in a single model.}
\label{fig:compare_DT_Metfaces}
\end{figure*}

\begin{figure*}[t!]
\centering
\includegraphics[width=0.82\linewidth]{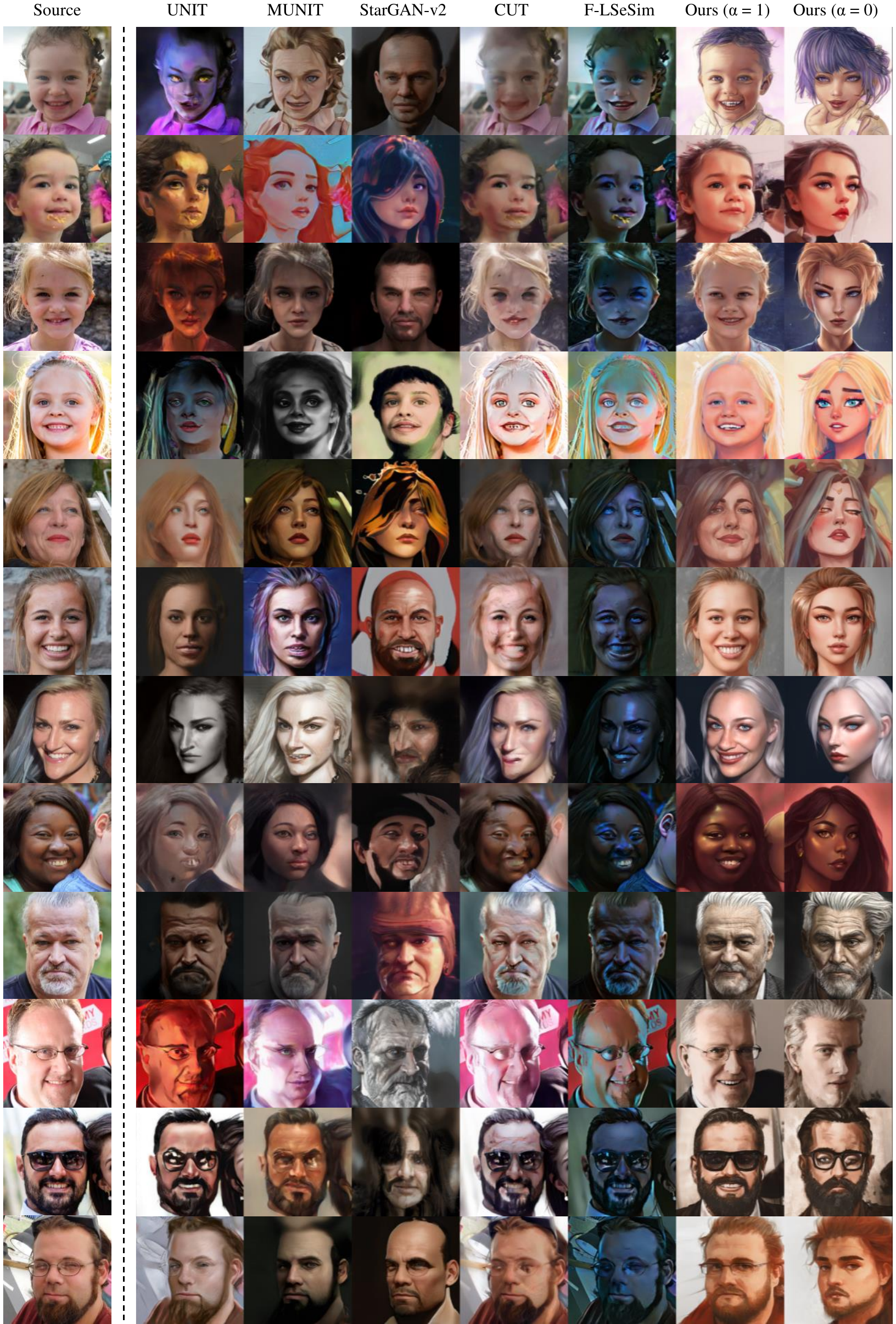}
   \caption{\textbf{[FFHQ $\rightarrow$ AAHQ]}
   Qualitative comparison on domain translation.
   Our method is not only qualitatively best, but also can control source features in a single model.}
\label{fig:compare_DT_AAHQ}
\end{figure*}

\begin{figure*}[t!]
\centering
\includegraphics[width=0.84\linewidth]{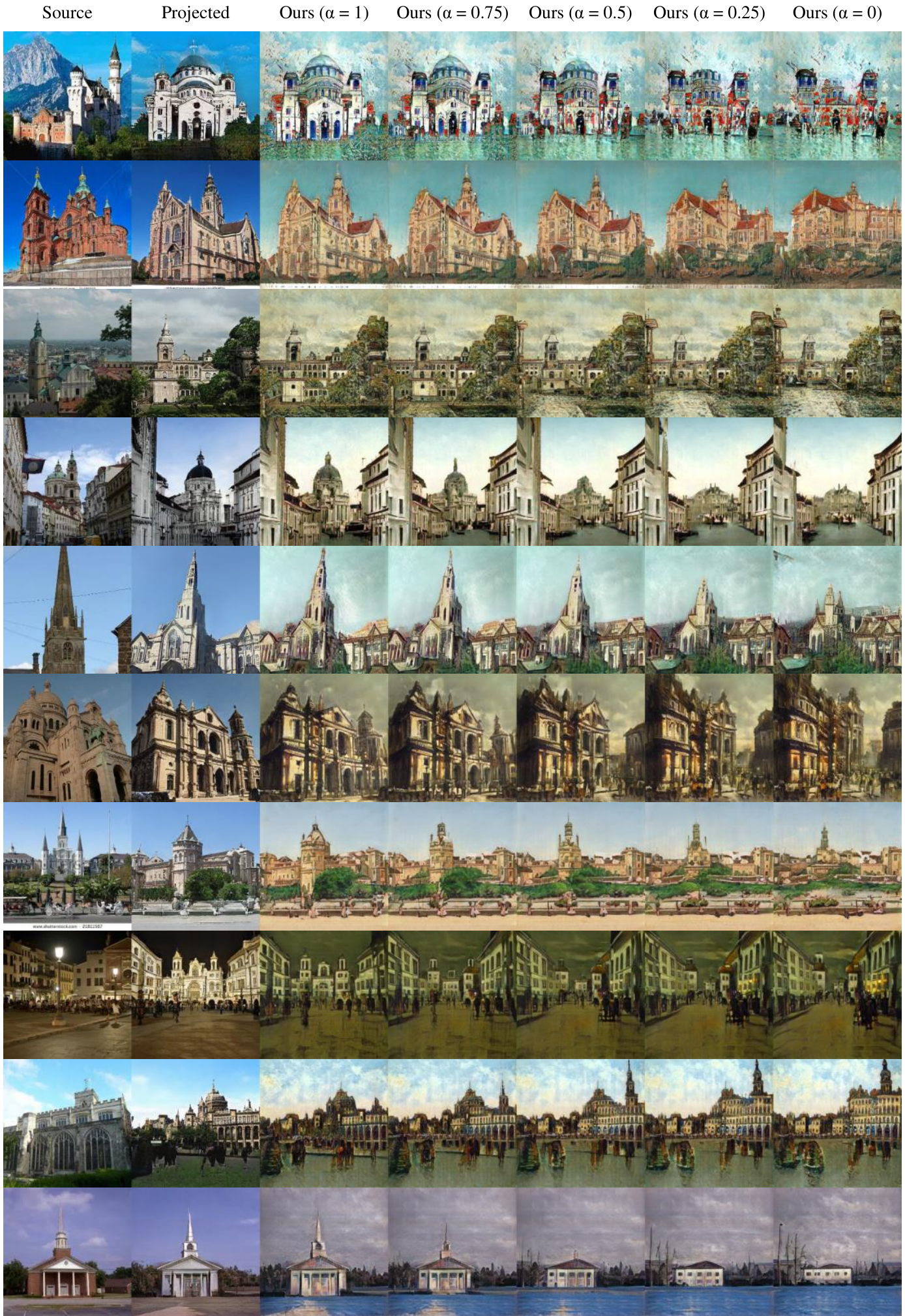}
   \caption{
   Domain translation results for Church $\rightarrow$ Cityscape. Incorrectly projected images cause our method to generate uncorrelated target images with source.}
\label{fig:DT_church_fail}
\end{figure*}

\begin{figure*}[t!]
\centering
\includegraphics[width=0.8\linewidth]{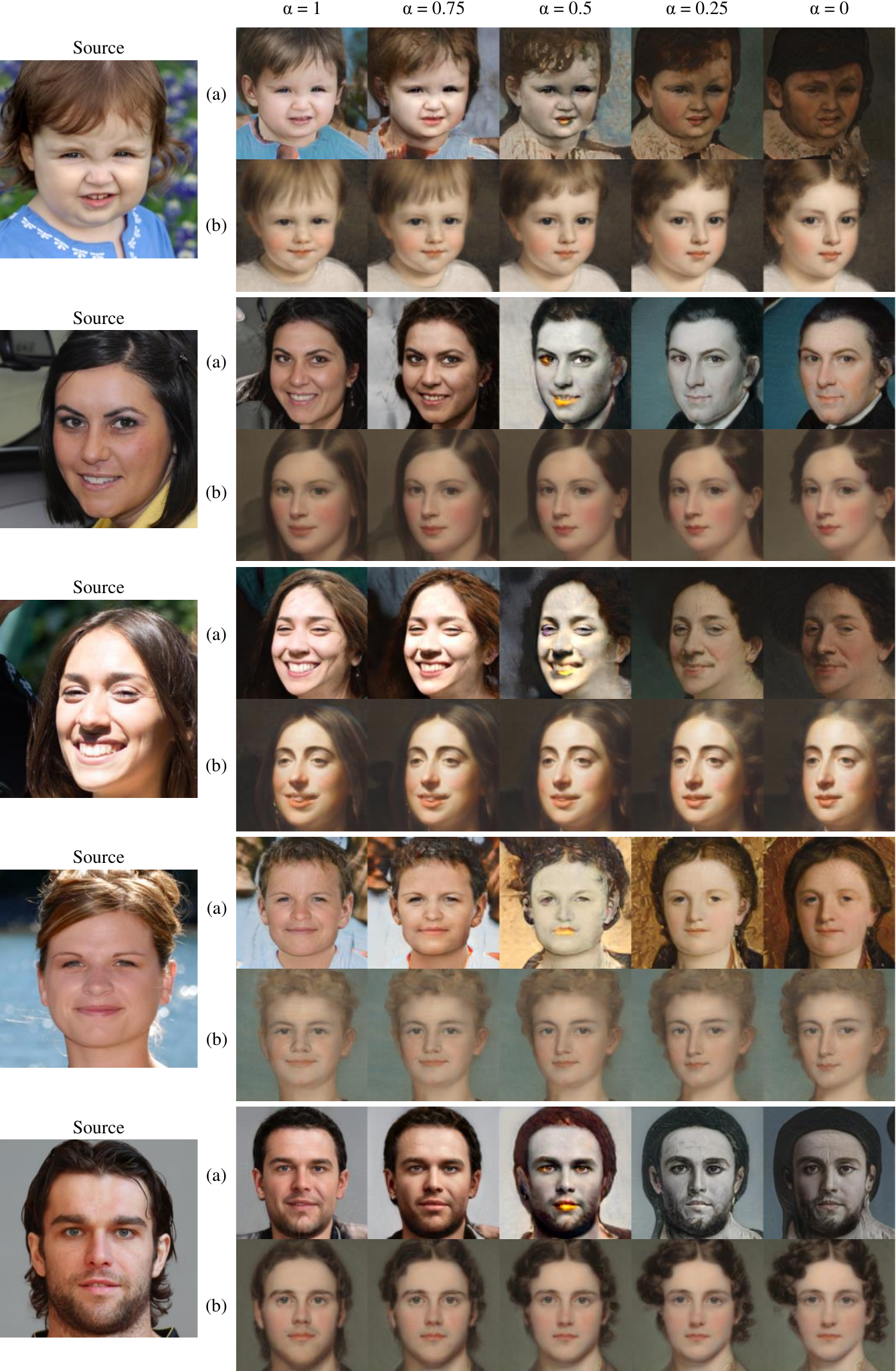}
   \caption{\textbf{[FFHQ $\rightarrow$ MetFaces]}
   Qualitative comparison on interpolation between source and target features: (a) SmoothingLatentSpace \cite{Liu_2021_CVPR}, (b) ours.
   The interpolation weight $\alpha$ is presented above each column.
}
\label{fig:compare_DT_smoothing_metfaces}
\end{figure*}

\begin{figure*}[t!]
\centering
\includegraphics[width=0.8\linewidth]{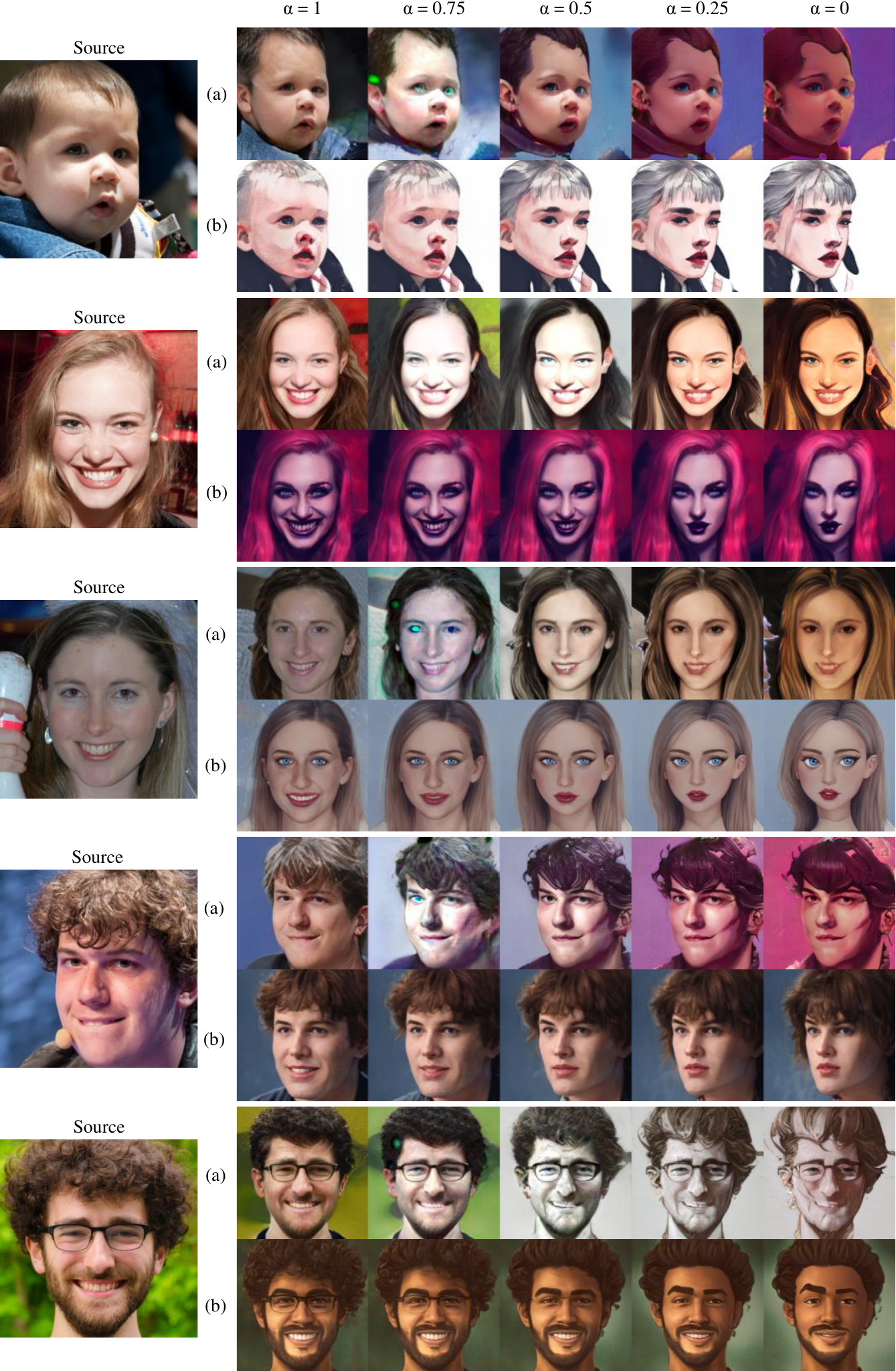}
   \caption{\textbf{[FFHQ $\rightarrow$ AAHQ]}
   Qualitative comparison on interpolation between source and target features: (a) SmoothingLatentSpace \cite{Liu_2021_CVPR}, (b) ours.
   The interpolation weight $\alpha$ is presented above each column.
}
\label{fig:compare_DT_smoothing_AAHQ}
\end{figure*}
\fi

\end{document}